\documentclass[runningheads]{llncs}
\usepackage{graphicx}
\usepackage[dvipsnames, svgnames, x11names]{xcolor}
\usepackage{tikz}
\usepackage{comment}
\usepackage{amsmath,amssymb} 
\usepackage{color}
\usepackage{graphicx}
\usepackage{subfigure}
\usepackage{amsmath}
\usepackage{amssymb}
\usepackage{booktabs}
\usepackage{float}
\usepackage{capt-of}
\usepackage{cuted}
\usepackage{comment}
\usepackage{multirow}
\usepackage{arydshln}
\usepackage{caption2}
\usepackage{mathrsfs}

\usepackage[accsupp]{axessibility}  

\usepackage[pagebackref,breaklinks,colorlinks]{hyperref}
\usepackage[capitalize]{cleveref}
\newcommand{\etal}{et al.}
\newcommand{\ie}{i.e.}
\newcommand{\eg}{e.g.}

\definecolor{myblue}{RGB}{10,50,90}
\usepackage{hyperref}
\hypersetup{
    colorlinks=true,
    linkcolor=red,
    filecolor=blue,      
    urlcolor=blue,
    citecolor=myblue,
}
\usepackage{marvosym}

\begin{document}
\pagestyle{headings}
\mainmatter
\def\ECCVSubNumber{4080}  

\title{Towards Efficient and Scale-Robust Ultra-High-Definition Image Demoiréing} 

\titlerunning{ECCV-22 submission ID \ECCVSubNumber} 
\authorrunning{ECCV-22 submission ID \ECCVSubNumber} 
\author{Anonymous ECCV submission}
\institute{Paper ID \ECCVSubNumber}
\titlerunning{Towards Efficient and Scale-Robust UHD Image Demoiréing}
%
\renewcommand{\thefootnote}{\textrm{\Letter}} 

\author{Xin Yu\inst{1} \and
Peng Dai\inst{1} \and
Wenbo Li\inst{2} \and
Lan Ma\inst{3} \and
\\
Jiajun Shen\inst{3} \and
Jia Li\inst{4}\and
Xiaojuan Qi\inst{1\textrm{\Letter}}}
\authorrunning{X. Yu et al.}
%
\institute{The University of Hong Kong \and
The Chinese University of Hong Kong \and
TCL AI Lab \and
Sun Yat-sen University}
\maketitle

\footnotetext[1]{indicates the corresponding author.} 

\begin{abstract}
  With the rapid development of mobile devices, modern widely-used mobile phones typically allow users to capture 4K resolution (\ie, ultra-high-definition)
  images. However, for image demoiréing, a challenging task in low-level vision, existing works are generally carried out on low-resolution or synthetic images. Hence, the effectiveness of these methods on 4K resolution images is still unknown. In this paper, we explore moiré pattern removal for ultra-high-definition images. To this end, we propose the first ultra-high-definition demoiréing dataset (UHDM), which contains 5,000 real-world 4K resolution image pairs, and conduct a benchmark study on current state-of-the-art methods. Further, 
  we present an efficient baseline model ESDNet for tackling 4K moiré images, wherein we build a semantic-aligned scale-aware module to address the scale variation of moiré patterns. Extensive experiments manifest the effectiveness of our approach, which outperforms state-of-the-art methods by a large margin while being much more lightweight. Code and dataset are available at \href{https://xinyu-andy.github.io/uhdm-page}{https://xinyu-andy.github.io/uhdm-page}.

\end{abstract}
  
\section{Introduction}
\label{sec:intro}

When photographing the contents displayed on the digital screen, an inevitable frequency aliasing between the camera's color filter array (CFA) and the screen's LCD subpixel widely exists. The captured images are thus mixed with colorful stripes, named moiré patterns, which severely degrade the perceptual quality of images. Currently, efficiently removing moiré patterns from a single moiré image is still challenging and receives growing attention from the research community.  

Recently, several image demoiréing methods~\cite{he2020fhde,zheng2020image,he2019mop,sun2018moire,liu2020wavelet,cheng2019multi,liu2018demoir,yuan2019aim} have been proposed, yielding a plethora of dedicated designs such as moiré pattern classification~\cite{he2019mop}, frequency domain modeling \cite{liu2020wavelet,zheng2020image}, and multi-stage framework \cite{he2020fhde}. Apart from FHDe$^2$Net \cite{he2020fhde} which is specially designed for high-definition images, most of the research efforts have been devoted to studying low-resolution images \cite{sun2018moire} or synthetic images \cite{yuan2019aim}.
However, the fast development of mobile devices enables modern mobile phones to capture ultra-high-definition images, so it is more practical to conduct research on 4K image demoiréing for real applications. Unfortunately, the highest resolution in current public demoiréing datasets (see \cref{tab:datasets}) is 1080p \cite{he2020fhde} ($1920\times 1080$). Whether methods investigated on such datasets can be trivially transferred into the 4K scenario is still unknown due to the data distribution change and dramatically increased computational cost. 

Under this circumstance, we explore the more practical yet more challenging demoiréing scenario, \ie, ultra-high-definition image demoiréing.  To evaluate the demoiréing methods in this scenario, we build the first large-scale real-world ultra-high-definition demoiréing dataset (UHDM), which consists of $4,500$ training image pairs and $500$ testing image pairs with diverse scenes (see \cref{fig:dataset}).

\noindent\textbf{Benchmark study and limitation analysis:} Based upon our dataset, we conduct a benchmark study on state-of-the-art methods \cite{he2020fhde,zheng2020image,he2019mop,sun2018moire,liu2020wavelet,cheng2019multi}. 
Our empirical study reveals that most methods \cite{sun2018moire,cheng2019multi,zheng2020image} struggle to  remove moiré patterns with a much wider range of scales in 4K images while simultaneously tolerating the growing demands for computational cost (see \cref{fig:cost}) or fine image detail \cite{he2020fhde} (see \cref{fig:fhd}).
We attribute their deficiencies to the lack of an effective multi-scale feature extraction strategy.
Concretely, existing methods attempting to address the scale challenge can be coarsely categorized into two lines of research. One line of research develops multi-stage models, such as  FHDe$^2$Net~\cite{he2020fhde}, to process large moiré patterns at a low-resolution stage and then refines the textures at a high-resolution stage, which however incurs huge computational cost when applied to 4K images (see \cref{fig:cost}: FHDe$^2$Net). Another line of research utilizes features from different depths of a network to build multi-scale representations, in which the most representative work~\cite{zheng2020image} achieves a better trade-off between accuracy and efficiency (see \cref{fig:cost}: MBCNN), yet still cannot be generally scale-robust (see \cref{fig:fhd} and \cref{fig:visual}). We note that the extracted multi-scale features are from different semantic levels which may result in misaligned features when fused together, potentially limiting its capabilities.
Detailed study and analysis are unfolded in  \cref{sec:challenge}.

To this end, inspired by HRNet~\cite{wang2020deep},
we propose a plug-and-play semantic-aligned scale-aware module (SAM) to boost the network's capability in handling moiré patterns with diverse scales without incurring too much computational cost, serving as a supplement to existing methods. Specifically, SAM incorporates a pyramid context extraction module to effectively and efficiently extract multi-scale features aligned at the same semantic level. Further, a cross-scale dynamic fusion module is developed to selectively fuse multi-scale features where the fusion weights are learned and dynamically adapted to individual images.
  
Equipped with SAM, we develop an efficient and scale-robust network for 4K image demoiréing, named ESDNet. ESDNet adopts a simple encoder-decoder network with skip-connections as its backbone and stacks SAM at different semantic levels to boost the model's capability in addressing scale variations of 4K moiré images. ESDNet is easy to implement while achieving state-of-the-art performance (see \cref{fig:visual} and \cref{tab:FHDMi}) on the challenging ultra-high-definition image demoiréing dataset and three other public demoiréing datasets \cite{he2020fhde,yuan2019aim,sun2018moire}.
In particular, ESDNet exceeds multi-stage high-resolution method FHDe$^2$Net, \textbf{1.8dB} in terms of PSNR while being \textbf{300$\times$ faster (5.620s vs 0.017s)} 
in the UHDM dataset. Our major contributions are summarized as follows:
\begin{itemize}
  \item We are the first to explore the ultra-high-definition image demoiréing problem, which is more practical yet more challenging. To this end, we build a large-scale real-world 4K resolution demoiréing dataset UHDM. 
  \item We conduct a benchmark study for the existing state-of-the-art methods on this dataset, summarizing several challenges and analyses. Motivated by these analyses, we propose an efficient baseline model ESDNet for ultra-high-definition image demoiréing.
  \item Our ESDNet achieves state-of-the-art results on the UHDM dataset and three other public demoiréing datasets, in terms of quantitative evaluation and qualitative comparisons. Moreover, ESDNet is lightweight and can process standard 4K ($3840\times 2160$) resolution images at 60 fps.
\end{itemize}


\section{Related Work} 
\subsubsection{Image demoiréing:} To remove moiré patterns caused by the frequency aliasing,  
Liu \etal~\cite{liu2018demoir} propose a synthetic dataset by simulating the camera imaging process and develop a GAN-based~\cite{goodfellow2014generative} framework. Further, a large-scale synthetic dataset \cite{yuan2019aim} is proposed and promotes many follow-up works~\cite{zheng2020image,cheng2019multi,yuan2019aim}.
However, it is difficult for models trained on synthetic data to handle real-world scenarios due to the sim-to-real gap. 
For real-world image demoiréing, Sun \etal~\cite{sun2018moire} propose the first real-world moiré image dataset (\ie, TIP2018) and develop a multi-scale network (DMCNN).
To distinguish different types of moiré patterns, He \etal~\cite{he2019mop} manually annotate moiré images with category labels to train a moiré pattern classification model. 
Frequency domain methods~\cite{liu2020wavelet,zheng2020image} are also studied for moiré removal. 
To deal with high-resolution images, He \etal~\cite{he2020fhde} construct a high-definition dataset FHDMi and develop the multi-stage framework FHDe$^2$Net. 
Although significant progress has been achieved, the above methods either cannot achieve satisfactory results \cite{zheng2020image,he2019mop,sun2018moire,cheng2019multi} or suffer from heavy computational cost \cite{zheng2020image,he2020fhde,he2019mop,cheng2019multi}. More importantly, the highest resolution of existing image demoiréing datasets is FHDMi~\cite{he2020fhde} with 1080p resolution, which is not suitable for practical use considering the ultra-high-definition (4K) images captured by current mobile cameras. We focus on developing a lightweight model that can process ultra-high-definition images.

\subsubsection{Image restoration:} To this point, plenty of learning-based image restoration models have been proposed. For instance, residual learning~\cite{he2016deep} and dense connection~\cite{huang2017densely} are widely used to develop very deep neural networks for different low-level vision tasks~\cite{zhang2017beyond,anwar2020densely,lim2017enhanced,kim2016accurate,zhang2018residual}. In order to capture multi-scale information, encoder-decoder~\cite{ronneberger2015u} structures or hierarchical architectures are frequently exploited in image restoration tasks~\cite{zhang2019deep,zamir2021multi,gao2019dynamic}. Inspired by iterative solvers, some methods utilize recurrent structures~\cite{gao2019dynamic,tao2018scale} to gradually recover images while reducing the number of parameters. 
To preserve structural and semantic information, 
many works~\cite{xie2019image,liu2018image,song2018contextual,yang2017high,suvorov2021resolution,wang2020vcnet} adopt the perceptual loss~\cite{johnson2016perceptual} or generative loss~\cite{goodfellow2014generative,gulrajani2017improved,arjovsky2017wasserstein} to guide the training procedure. 
In our work, we also take advantage of the well-designed dense blocks for efficient feature reuse and the perceptual loss for semantically guided optimization.

\subsubsection{Multi-scale network:} The multi-scale network has been widely adopted in various tasks~\cite{wang2020deep,chen2018learning,zhou2018stereo,yeh2016semantic,chen2017photographic} due to its ability to leverage features with different receptive fields. U-Net~\cite{ronneberger2015u}, as one representative multi-scale network, extracts multi-scale information using an encoder-decoder structure, and enhances features in decoder with skip-connections. To preserve the high-resolution representation, the full resolution residual network~\cite{pohlen2017full} extends the U-Net by introducing an extra stream containing information of the full resolution, and similar operations can be found in the HRNet~\cite{wang2020deep}. Considering that the extracted multi-scale features have different semantic meanings, the question of how to fuse features with different meanings is also important and has been widely studied in many works~\cite{cai2016unified,chen2017deeplab,chen2018cascaded}. In this work, we design a semantic-aligned scale-aware module to handle moiré patterns with diverse scales without incurring too great a computational cost, which renders our method highly practical for 4K images.

\section{UHDM Dataset}
\label{sec:dataset}
We study ultra-high-definition image demoiréing, which has more practical applications. For the training of 4K demoiréing models and the evaluation of existing methods, we collect a large-scale ultra-high-definition demoiréing dataset (UHDM). Dataset collection and benchmark study are elaborated upon below.

\subsection{Data Collection and Selection}
\begin{figure}[t]\centering
  \includegraphics[width=1\linewidth]{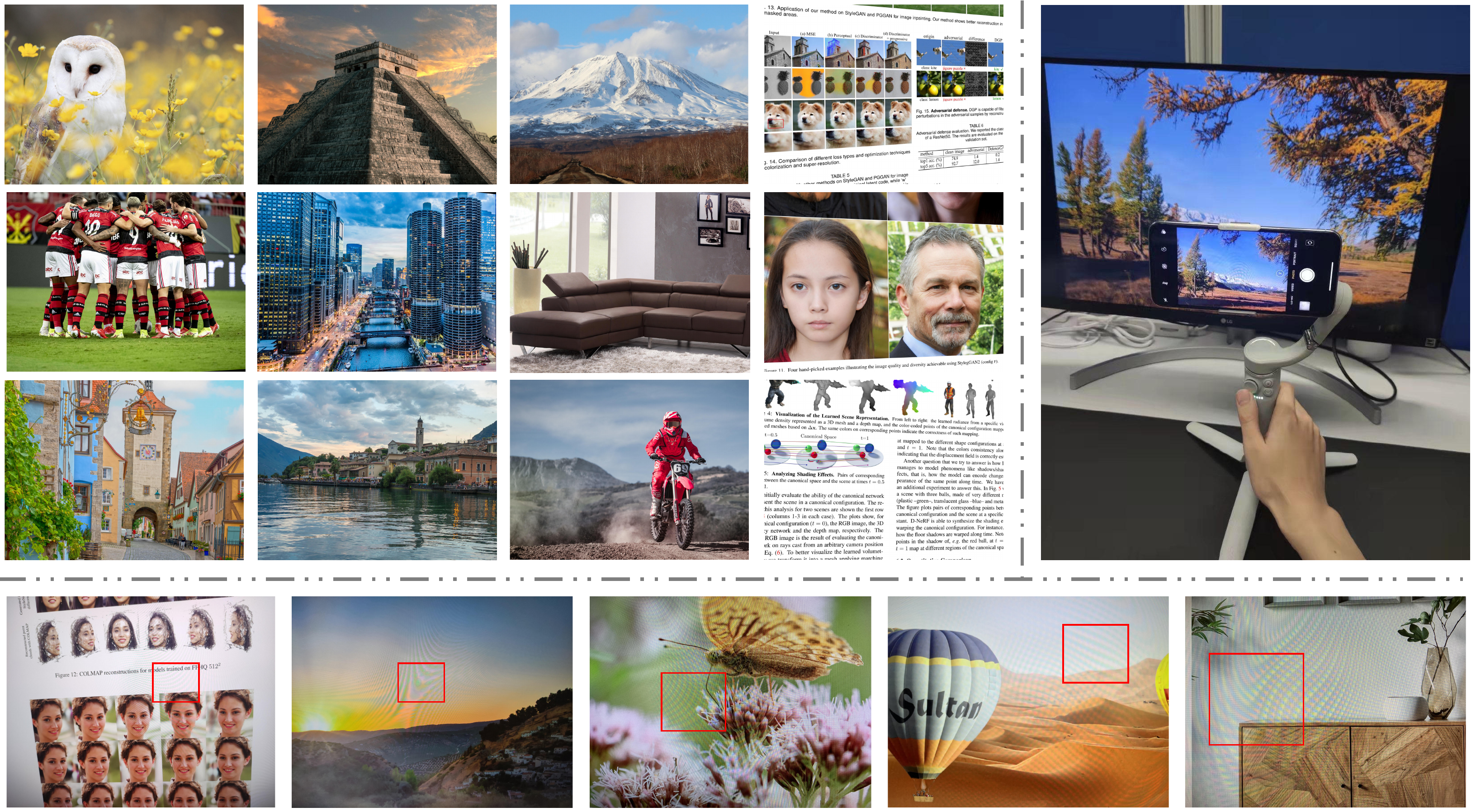}
  \caption{Upper left: Our dataset contains diversified scenarios. Upper right: we capture the moiré image with a DJI OM 5 smartphone gimbal. Lower: moiré images in our dataset show a wide range of scale variations}
  \label{fig:dataset} 
\end{figure}

To obtain the real-world 4K image pairs, we first collect high-quality images with resolutions ranging from 4K to 8K from the Internet.
We note that Internet resources lack document scenes, which also constitute a vital application scenario (\eg, slides, papers), so we manually generate high-quality text images and make sure they maintain 3000 dpi (Dots Per Inch). 
Finally, the collected moiré-free images cover a wide range of scenes (see \cref{fig:dataset}), such as landscapes, sports, video clips, and documents. 
Given these high-quality images, we generate diverse real-world moiré patterns elaborated upon below.

First, to produce realistic moiré images and ease the difficulties of calibrations, 
we shoot the clean pictures displayed on the screen with a camera phone fixed on a DJI OM 5 smartphone gimbal, which allows us to conveniently and flexibly adjust the camera view through its control button, as shown in \cref{fig:dataset}. 
Second, we note that the characteristics of moiré patterns highly are highly dependent upon the geometric relationship between the screen and the camera (see supplement for more details). 
Therefore, during the capturing process, we continuously adjust the viewpoint every ten shots to produce diverse moiré patterns. Third, we adopt multiple $<\text{mobile phone}, \text{screen}>$ (\ie, three mobile phones and three digital screens, see supplement for more details) combinations to cover various device pairs, since they will also have an impact on the styles of moiré patterns.
Finally, to obtain aligned pairs, we utilize RANSAC algorithm \cite{vedaldi2010vlfeat} to estimate the homography matrix between the original high-quality image and the captured moiré screen image. Since it is difficult to ensure accurate pixel-wise calibration due to the camera's internal nonlinear distortions and perturbations of moiré artifacts, manual selection is performed to rule out severely misaligned image pairs, thereby ensuring quality.

Our dataset contains $5,000$ image pairs in total. We randomly split them into $4,500$ for training and $500$ for validation. As we collect moiré images using various mobile phones, the resolution can either be $4032 \times 3024$ or $4624 \times 3472$. 
Comparisons with other existing datasets are shown in \cref{tab:datasets}, and the characteristics of our dataset are summarized as below.

\begin{itemize}
\item \textbf{Ultra-high resolution} 
UHDM is the first 4K resolution demoiréing dataset, consisting of 5,000 image pairs in total. 
\item \textbf{Diverse image scenes}
The dataset includes diverse scenes, such as landscapes, sports, video clips, and documents.
\item \textbf{Real-world capture settings} The moiré images are generated following practical routines, with different device combinations and viewpoints to produce diverse moiré patterns.
\end{itemize}

\begin{table*}[t]
   \centering
   \caption{Comparisons of different demoiréing datasets; our dataset is the first ultra-high-definition dataset (``London's Buildings'' is not available currently)}
   \renewcommand\tabcolsep{5.0pt}
   \resizebox{10cm}{!}
   {
   \begin{tabular}{c|cccc}
   \toprule[1.2pt]
   Dataset &Avg. Resolution &Size &Diversity &Real-world \\
   \hline 
   TIP18 \cite{sun2018moire} &$256\times 256$ &135,000 &No text scenes &$\checkmark$\\
   
   LCDMoiré \cite{yuan2019aim} &$1024\times 1024$ &10,200 &Only text scenes &$\times$\\
   
   FHDMi \cite{he2020fhde} &$1920\times 1080$ &12,000 &Diverse scenes &$\checkmark$\\
   
   London's Buildings \cite{liu2020wavelet} &$2100\times 1700$ &460 &Only urban scenes &$\checkmark$\\
   \midrule[0.8pt]
   \textbf{UHDM} &$\mathbf{4328\times3248}$ &\textbf{5,000} &\textbf{Diverse scenes} &$\mathbf{\checkmark}$\\
   \bottomrule[1.2pt]
   \end{tabular}
    }
\label{tab:datasets}
\end{table*}

\subsection{Benchmark Study on 4K Demoiréing}
\label{sec:challenge}

\begin{figure}[t]\centering
  \includegraphics[width=0.96\linewidth]{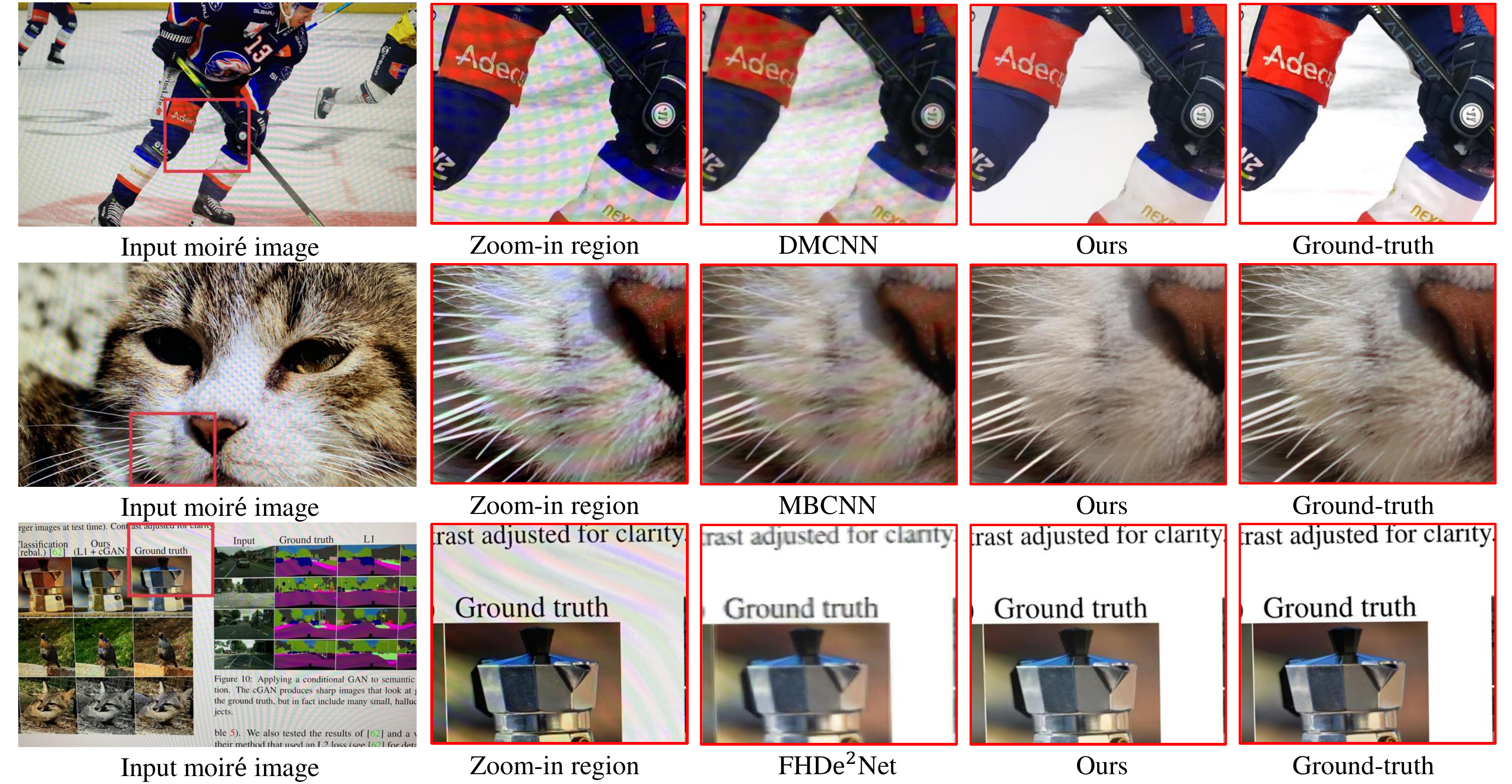}
  \caption{Limitations of current methods: they are often unable to remove the moiré pattern with a wider scale range or lose high-frequency details}
  \label{fig:fhd} 
\end{figure}

\begin{figure}[t]\centering
  \includegraphics[width=0.68\linewidth]{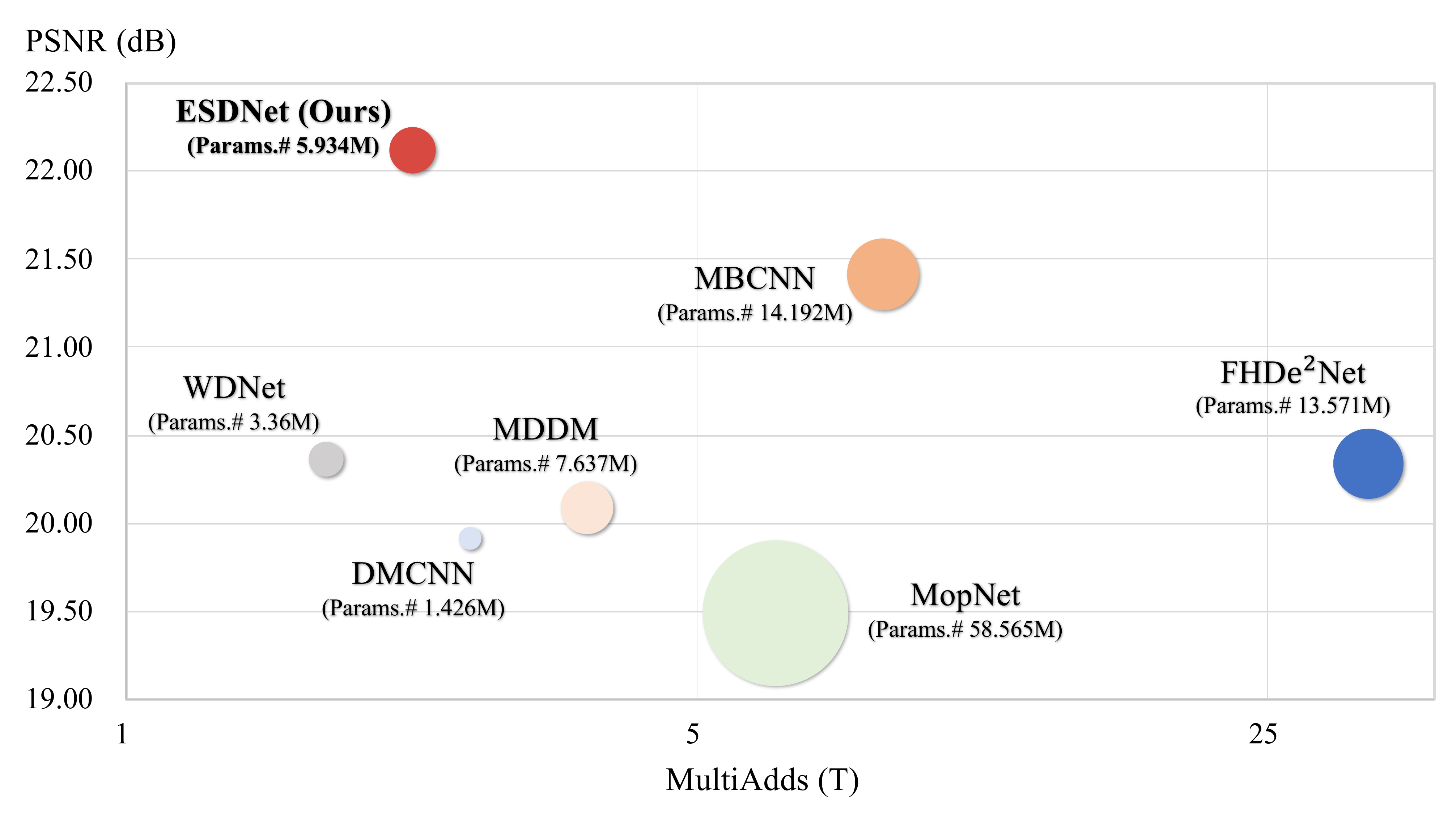}
  \caption{Comparisons of computational cost of different methods: The x-axis and the y-axis denote the MultiAdds (T) and PSNR (dB). The number of parameters is expressed by the area of the circle}
  \label{fig:cost} 
\end{figure}

As the image resolution is increased to the 4K resolution, the scale of moiré patterns has a very wide range, from very large moiré patterns to small ones (see \cref{fig:dataset}). This poses a major challenge to demoiréing methods as they are required to be scale-robust. Furthermore, increased image resolution also leads to dramatically increased computational cost and high requirements of detail restoration/preservation.
Here, we carry out a benchmark study on the existing state-of-the-art methods
\cite{zheng2020image,sun2018moire,he2019mop,he2020fhde,liu2020wavelet,cheng2019multi} on our 4K demoiréing dataset to evaluate their effectiveness. Main results are summarized in \cref{fig:fhd} and \cref{fig:cost}: existing methods are mostly not capable of achieving a good balance of accuracy and computational efficiency.
More detailed results are shown in \cref{sec:ex}.

\noindent{\textbf{Analysis and discussions:}} Although existing methods also attempt to address the scale challenge by developing multi-scale strategies, they still have several deficiencies regarding computational efficiency and restoration quality when applied to 4K high-resolution images (see \cref{fig:fhd}). One line of methods, such as DMCNN~\cite{sun2018moire} and MDDM \cite{cheng2019multi}, fuses multi-scale features harvested from multi-resolution inputs only at the output stage, which potentially prohibits the intermediate features from interacting with and refining each other, leading to sub-optimal results, \ie, significantly sacrificing accuracy on 4K image demoiréing despite being lightweight (see \cref{fig:cost} and  \cref{fig:fhd}). 
Another line of methods, such as MBCNN~\cite{zheng2020image}, exploits multi-scale features at different network depths following a U-Net-like architecture. Compared with other existing methods, although it achieves the best trade-off between accuracy and efficiency, it still suffers from moiré patterns with a wide-scale range (the second row of \cref{fig:fhd} and \cref{fig:visual}). One possible issue is that the combined multi-scale features come from different semantic levels~\cite{wang2020deep}, prohibiting a specific feature level to harvest multi-resolution representations \cite{wang2020deep}, which could also be an important cue for image demoiréing.
On the other hand, FHDe$^2$Net~\cite{he2020fhde} designs a coarse-to-fine two-stage model to simultaneously address the scale and detail challenge. It suffers, however, from heavy computational cost when applied to 4K images (see \cref{fig:cost}) yet is still not sufficient to remove moiré patterns (see \cref{fig:visual}) or recover fine image detail (see  \cref{fig:fhd} and \cref{fig:visual}).

\section{Proposed Method}\label{sec:method}

Motivated by observations in \cref{sec:challenge}, we introduce a baseline approach to advance 4K resolution image demoiréing, aimed towards a more scale-robust and efficient model. In the following, we first present an overview of our pipeline and then elaborate on our core semantic-aligned scale-aware module (SAM).

 
\begin{figure*}[t]
    \includegraphics[width=1\textwidth]{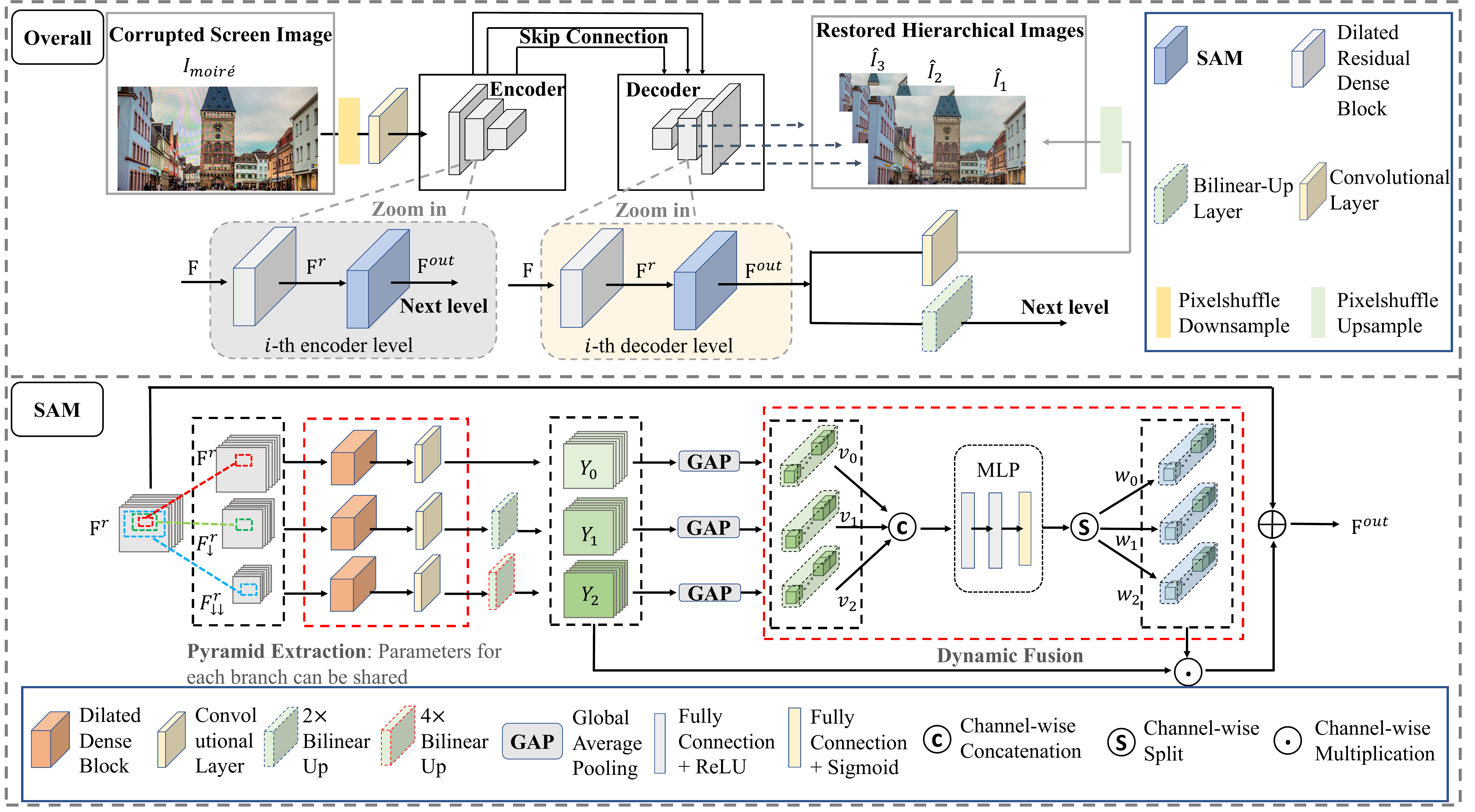}
    \caption{The pipeline of our ESDNet and the proposed semantic-aligned scale-aware module (SAM) 
    }
    \label{fig:ESDNet}
\end{figure*}

\subsection{Pipeline}
The overall architecture is shown in \cref{fig:ESDNet}, where a pre-processing head is utilized to enlarge the receptive field, followed by an encoder-decoder architecture for image demoiréing.
The pre-processing head adopts pixel shuffle \cite{shi2016real} to downsample the image by two times and a $5\times 5$ convolution layer to further extract low-level features. 
Then, the extracted low-level features are fed into an encoder-decoder backbone architecture that consists of three downsampling and upsampling levels. Note that the encoder and decoder are connected via skip-connections to allow features containing high-resolution information to facilitate the restoration of corresponding moiré-free images. At each decoder level, the network would produce intermediate results through a convolution layer and a pixelshuffle upsampling operation (see the upper part of \cref{fig:ESDNet}), which are also supervised by the ground-truth, serving the purpose of deep supervision to facilitate training.
Specifically, each encoder or decoder level (see \cref{fig:ESDNet}) contains a dilated residual dense block  \cite{zhang2018residual,huang2017densely,he2016deep,yu2015multi} for refining the input features (as detailed below) and a proposed semantic-aligned multi-scale module (SAM) for extracting and dynamically fusing multi-scale features at the same semantic level (as elaborated in \cref{sec:MCRB}).

\noindent{\textbf{Dilated residual dense block:}}
For each level $i\in\{1,2,3,4,5,6\}$ (\ie, three encoder levels and three decoder levels), the input feature $F_i$ first goes through a convolutional block, \ie, dilated residual dense block, for refining input features. It incorporates the residual dense block (RDB) \cite{zhang2018residual,huang2017densely,he2016deep} and dilated convolution layers \cite{yu2015multi} to process the input features and output refined ones. Specifically, given an input feature $F_i^0$ to the $i$-th level encoder or decoder, the cascaded local features from each layer inside the block can be formulated as Eq.~\eqref{eq:denseblock}:

\begin{equation}
F_{i}^l=C^l([F_{i}^0, F_{i}^1,..., F_{i}^{l-1}]), (l=1,2,...,L) \label{eq:denseblock}
\end{equation}
where $[F_{i}^0, F_{i}^1,..., F_{i}^{l-1}]$ denotes the concatenation of all intermediate features inside the block before layer $l$, and $C^l$ is the operator to process the concatenated features, consisting of a $3\times 3$ Conv with dilated rate $d^l$ and a rectified linear unit (ReLU). 
After that,
we apply a $1\times 1$ convolution to keep the output channel number the same
as that of $F_{i}^0$. Finally, we exploit the residual connection to produce the refined feature representation $F_i^r$, formulated as Eq.\eqref{eq:refine}:
\begin{equation}
F_i^r = F_{i}^0+\text{Conv$_{1\times 1}$}(F_{i}^L). \label{eq:refine}
\end{equation}

\noindent The refined feature representation $F_i^r$ is then fed to our proposed SAM for semantic-aligned multi-scale feature extraction.

\subsection{Semantic-Aligned Scale-Aware Module}\label{sec:MCRB}
Given the input feature $F_i^r$, the SAM is intended to extract multi-scale features within the same semantic level $i$ and allow them to interact and be dynamically fused, significantly improving the model's ability to handle moiré patterns with a wide range of scales.
As demonstrated in \cref{tab:ablation}, SAM enables us to develop a lightweight network while still being more effective in comparison with existing methods. In the following, we detail the design of SAM which encompasses two major modules: pyramid feature extraction and cross-scale dynamic fusion.

\subsubsection{Pyramid context extraction:}
Given an input feature map $F^r\in\mathbb{R}^{H \times W \times C}$ (we simplify $F_i^r$ by $F^r$ in the following discussion), we first produce pyramid input features $F^r\in\mathbb{R}^{H \times W \times C}, F_{\downarrow}^r\in\mathbb{R}^{\frac{H}{2} \times \frac{W}{2} \times C}$ and $ F_{\downarrow \downarrow}^r\in \mathbb{R}^{\frac{H}{4} \times \frac{W}{4} \times C}$ through bilinear interpolation, 
then feed them into a corresponding convolutional branch with five convolution layers to yield  pyramid outputs $Y_0, Y_1, Y_2$ (see the lower part of \cref{fig:ESDNet}):
\begin{equation}
Y_{0}=E_{0}(F^r), \quad Y_{1}=E_{1}\left(F_{\downarrow}^r\right), \quad Y_{2}=E_{2}\left(F_{\downarrow \downarrow}^r\right),
\end{equation}
where we build $E_{0}, E_{1}$, and $E_{2}$ via the dilated dense block, followed by a $1\times 1$ convolution layer. In addition, the up-sampling operations will be  performed in $E_1, E_2$ to align the size of three outputs, \ie, $Y_i\in \mathbb{R}^{H \times W \times C}, (i=0,1,2)$.

Note that, as the internal architectures of $E_0, E_1$, and $E_2$ are identical, their corresponding learnable parameters can be shared to lower the cost of parameter number. In fact, as proven in \cref{sec:ex}, the improvement primarily comes from the pyramid architecture instead of additional parameters.

\subsubsection{Cross-scale dynamic fusion:}
Given the pyramid features $Y_0, Y_1, Y_2$, the cross-scale dynamic fusion module fuses them together to produce fused multi-scale features for the next level to process. The insight for this module is that scale of moiré patterns vary from image to image and thus the importance of different scale features would also vary across images. Therefore, we develop the following cross-scale dynamic fusion module to make the fusion process dynamically adjusted and adapted to each image. Specifically, we learn dynamic weights to fuse $Y_1, Y_2, Y_3$.

Given $Y_i\in\mathbb{R}^{H \times W \times C}  (i=0,1,2)$, we first apply global average pooling in the spatial dimension of each feature map to obtain the 1D global feature $v_i\in\mathbb{R}^{C}$ for each scale $i$ following Eq. \eqref{eq:global}.

\begin{equation}
v_i =\frac{1}{H \times W} \sum_{s=1}^{H} \sum_{t=1}^{W}Y_i(s, t) \label{eq:global}
\end{equation}
Then, we concatenate them along the channel dimension and learn the dynamic weights through an MLP module as:

\begin{equation}
\left[w_{0}, w_{1}, w_{2}\right]=\text{MLP}(\left[v_{0}, v_{1}, v_{2}\right])
\end{equation}
where ``MLP'' consists of three fully connected layers and outputs $w_{0}, w_{1}, w_{2}\in \mathbb{R}^{{C}}$ to fuse $Y_1, Y_2, Y_3$ dynamically. Finally, with fusion weights, we channel-wisely fuse the pyramid features with the input-adaptive weights, and then add the input feature $F^r$ to get the final output of SAM:
\begin{equation}
F^{\text{out}} = F^r+ w_0\odot Y_0+ w_1\odot Y_1+ w_2\odot Y_2
\end{equation}

\noindent where $\odot$ denotes the channel-wise multiplication, and the output $F^{\text{out}}$ will go through the next level ($i\rightarrow i+1$) for further feature extraction and image reconstruction.

\subsubsection{Comparisons and analysis:}
Existing methods \cite{zheng2020image,liu2020wavelet} utilize features from different depths to obtain multi-scale representations. However, features at different depths have different levels of semantic information. Thus, they are incapable of representing multi-scale information at the same semantic level, which might provide important cues for boosting the model's multi-scale modeling capabilities, as indicated in \cite{wang2020deep}.
We offer SAM as a supplement to existing methods as $Y_0, Y_1, Y_2$ include semantic-aligned information with different local receptive fields. The dynamic fusion methods further make the module adaptive to different images and boost its abilities. This strategy can also be treated as an implicit classifier compared with the explicit one in MopNet \cite{he2019mop}, which is more efficient and avoids the ambiguous hand-craft attribute definition. We include more detailed analysis in our supplementary file.

\subsection{Loss Function}

To boost optimization, we adopt the deep supervision strategy, which has been proven useful in \cite{zheng2020image}.
As shown in \cref{fig:ESDNet}, in each decoder level, the network will produce hierarchical predictions $\hat{I}_{1}, \hat{I}_{2}, \hat{I}_{3},$ which are also supervised by ground-truth images. 
We note that moiré patterns disrupt image structures since they generate new strip-shaped structures. Therefore, we adopt the perceptual loss~\cite{johnson2016perceptual} for feature-based supervision. 
At each level, we build our loss function by combining the pixel-wise $L_1$ loss and the feature-based perceptual loss $L_p$. Hence, the final loss function is formulated as:
\begin{equation}
\mathcal{L}_{total} = \sum_{i=1}^{3}\mathcal{L}_{1}(I_i, \hat{I}_{i})+\lambda\times\mathcal{L}_{p}(I_i, \hat{I}_{i})
\end{equation}
For the perceptual loss, we extract features from conv3\_3 (after ReLU) using a pre-trained VGG16~\cite{simonyan2014very} network and compute the $L_1$ distance in the feature space; we simply set $\lambda=1$ during training. We find that this perceptual loss is effective in removing moiré patterns. 

\section{Experiments}
\label{sec:ex}

\noindent\textbf{Datasets and metrics:}
We conduct experiments on the proposed UHDM dataset and three other public datasets: FHDMi~\cite{he2020fhde}, TIP2018~\cite{sun2018moire} and LCDMoiré~\cite{yuan2019aim}. 
In our UHDM dataset, we keep the original two resolutions (see \cref{sec:dataset}) and models are trained with cropped patches. During the evaluation phase, we do center crop from the original images to obtain test pairs with a resolution of $3840\times 2160$ (standard 4K size).
We adopt the widely used PSNR, SSIM~\cite{wang2004image} and LPIPS~\cite{zhang2018unreasonable} metrics for quantitative evaluation. It has been proven that LPIPS is more consistent with human perception and suitable for measuring demoiréing quality~\cite{he2020fhde}. 
Note that existing methods only report PSNR and SSIM on the TIP2018 and LCDMoiré, so we follow this setup for comparisons. 

\noindent\textbf{Implementation details:}
We implement our algorithm using PyTorch on an NVIDIA RTX 3090 GPU card. During training, we randomly crop a $768 \times 768$ patch from the ultra-high-definition images, and set the batch size to $2$. The model is trained for $150$ epochs and optimized by Adam \cite{kingma2014adam} with $\beta_{1}= 0.9 $ and $\beta_{2} = 0.999$. The learning rate is initially set to $0.0002$ and scheduled by cyclic cosine annealing \cite{loshchilov2016sgdr}. Details for implementations on other benchmarks are unfolded in the supplementary file. We also train other methods on our dataset faithfully and sufficiently and unfold details in the supplementary file.

\begin{table*}[t]
\caption{Quantitative comparisons between our model and state-of-the-art methods on four datasets. $(\uparrow)$ denotes the larger the better, and $(\downarrow)$ denotes the smaller the better. \textcolor{red}{Red}: best and \textcolor{blue}{Blue}: second-best}
\centering
\resizebox{\textwidth}{!}{
\begin{tabular}{c|c|ccccccc|cc}
\toprule[1.2pt]
Dataset &Metrics   &Input &DMCNN\cite{sun2018moire} &MDDM\cite{cheng2019multi} &WDNet\cite{liu2020wavelet} &MopNet\cite{he2019mop}  &MBCNN\cite{zheng2020image} &FHDe$^{2}$Net\cite{he2020fhde}   & ESDNet & ESDNet-L\\
\hline \multirow{3}{*}{UHDM}
&\text{PSNR}$\uparrow$ &17.117 &19.914 &20.088 &20.364 &19.489 &21.414 &20.338 &\textcolor{blue}{22.119} &\textcolor{red}{22.422} \\

&\text{SSIM}$\uparrow$ &0.5089 &0.7575 &0.7441 &0.6497 &0.7572 &0.7932 &0.7496 &\textcolor{blue}{0.7956} &\textcolor{red}{0.7985}  \\

&\text{LPIPS}$\downarrow$ &0.5314 &0.3764 &0.3409 &0.4882 &0.3857 &0.3318 &0.3519 &\textcolor{blue}{0.2551} &\textcolor{red}{0.2454}  \\

\hline \multirow{3}{*}{FHDMi}
&\text{PSNR}$\uparrow$ &17.974  & 21.538 & 20.831 &- & 22.756 & 22.309 & 22.930  & \textcolor{blue}{24.500} & \textcolor{red}{24.882} \\

&\text{SSIM}$\uparrow$ &0.7033 & 0.7727 & 0.7343 &- & 0.7958 & 0.8095 & 0.7885  & \textcolor{blue}{0.8351} & \textcolor{red}{0.8440} \\

&\text{LPIPS}$\downarrow$ &0.2837 & 0.2477 & 0.2515 &- & 0.1794 & 0.1980 & 0.1688  & \textcolor{blue}{0.1354} & \textcolor{red}{0.1301} \\

\hline \multirow{2}{*}{TIP2018}
&\text{PSNR}$\uparrow$    & 20.30 & 26.77 & - &28.08 & 27.75 & \textcolor{blue}{30.03} & 27.78  & 29.81 & \textcolor{red}{30.11}\\

&\text{SSIM}$\uparrow$    & 0.738 & 0.871 & - &0.904 & 0.895 & 0.893 & 0.896 & \textcolor{blue}{0.916} & \textcolor{red}{0.920} \\

\hline \multirow{2}{*}{LCDMoiré}
&\text{PSNR}$\uparrow$    & 10.44 & 35.48 & 42.49 &29.66 & - & 44.04 & 41.40  & \textcolor{blue}{44.83} & \textcolor{red}{45.34}\\

&\text{SSIM}$\uparrow$    & 0.5717 & 0.9785 & 0.9940 &0.9670 & - & 0.9948 & - & \textcolor{blue}{0.9963} & \textcolor{red}{0.9966} \\

\hline
\hline

- & Params (M) & - & 1.426 & 7.637 & 3.360 & 58.565 & 14.192 & 13.571 & 5.934 & 10.623\\

\bottomrule[1.2pt]
\end{tabular}
}

\label{tab:FHDMi}
\end{table*}


\begin{figure}[t]\centering
\includegraphics[width=1\linewidth]{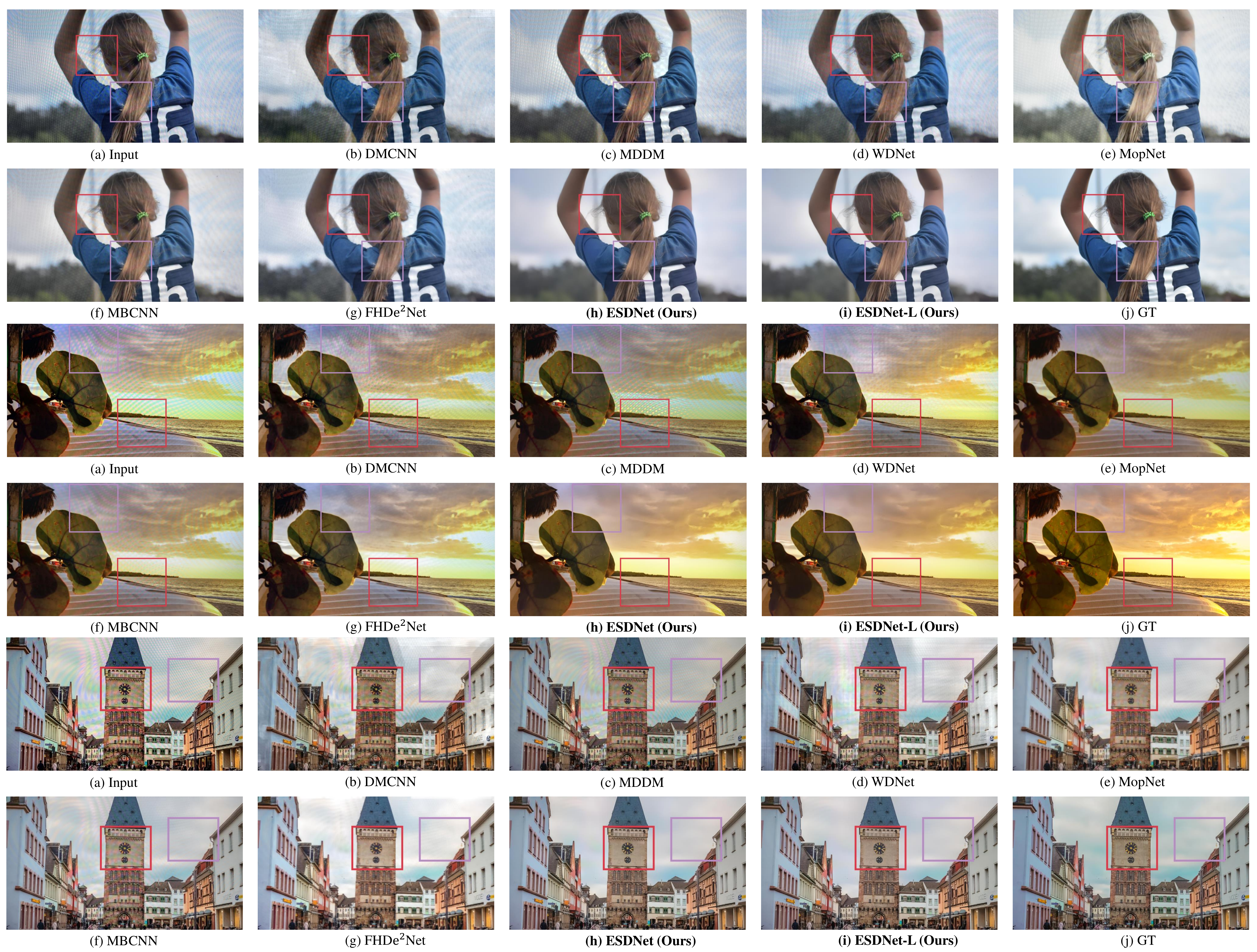}

\caption{Qualitative comparisons with state-of-the-art methods on the UHDM dataset. Please zoom in for a better view. More results are given in the supplementary file}
\label{fig:visual} 
\end{figure}


\subsection{Comparisons with State-of-the-Art Methods}
We provide two versions of our model: ESDNet and ESDNet-L.
ESDNet is the default lightweight model and ESDNet-L is a larger model, stacking one more SAM in each network level.

\noindent\textbf{Quantitative comparison:}
Table \ref{tab:FHDMi} shows quantitative performance of existing approaches. The proposed method achieves state-of-the-art results on all four datasets. Specifically, both of our two models outperform other methods by a large margin in the ultra-high-definition  UHDM dataset and high-definition FHDMi dataset, demonstrating the effectiveness of our method in high-resolution scenarios. It is worthwhile to note that our ESDNet, though possessing far fewer parameters, already shows competitive performance.

\noindent\textbf{Qualitative comparison:}
We present visual comparisons between our algorithm and existing methods in \cref{fig:visual}. Apparently, our method obtains more perceptually satisfactory results. In comparison, MDDM~\cite{cheng2019multi}, DMCNN~\cite{sun2018moire} and WDNet~\cite{liu2020wavelet} often fail to remove moiré patterns, while MBCNN~\cite{zheng2020image} and MopNet~\cite{he2019mop} cannot handle large-scale patterns well. Though 
performing better than other methods (except for ours), FHDe$^2$Net~\cite{he2020fhde}  usually suffers from severe loss of details. All these facts manifest the superiority of our method.

\noindent\textbf{Computational cost:} As shown in \cref{fig:cost}, our method strikes a sweet point of balancing the parameter number, computation cost (MACs), and demoiréing performance. Also, we test the inference speed of our method on an NVIDIA RTX 3090 GPU. Surprisingly, our ESDNet only needs 17ms (\ie, 60fps) to process a standard 4K resolution image, almost $300\times$ faster than FHDe$^2$Net. The competitive performance and low computational cost render our method highly practical in the 4K scenario.


\begin{table*}[t]
    \caption{Ablation study of the proposed SAM. ``A'' represents the baseline model. ``A$^+$'' denotes a stronger baseline which is of similar model capacity compared to our full model ``E''. ``B'' adds the pyramid context extraction with shared weights across all branches to ``A'' while ``D'' adopts adaptive weights. ``C'' and ``E'' add the cross-scale dynamic fusion based on ``B'' and ``D'', respectively}
   \centering
   \renewcommand\tabcolsep{5.0pt}
   \resizebox{10cm}{!}{
   \begin{tabular}{c|c|ccccc|c}
   \toprule[1.2pt]
   Dataset &Metrics   &A &A$^{+}$ &B &C &D &E\\
   \hline \multirow{3}{*}{UHDM}
   &\text{PSNR}$\uparrow$ &20.646 &20.860 &21.176 &21.958 &21.300  &\textbf{22.119}  \\
   
   &\text{SSIM}$\uparrow$ &0.7899 &0.7908 &0.7937 &0.7938 &0.7947  &\textbf{0.7956}  \\
   
   &\text{LPIPS}$\downarrow$ &0.2750 &0.2626 &0.2683 &0.2596 &0.2623  &\textbf{0.2551}  \\
   \hline
   &\text{Params} (M) &2.705 &5.978 &2.705 &3.014 &5.625  &5.934  \\
   \bottomrule[1.2pt]
\end{tabular}
}
\label{tab:ablation}
\end{table*} 

\begin{figure}[t]\centering
    \includegraphics[width=1\linewidth]{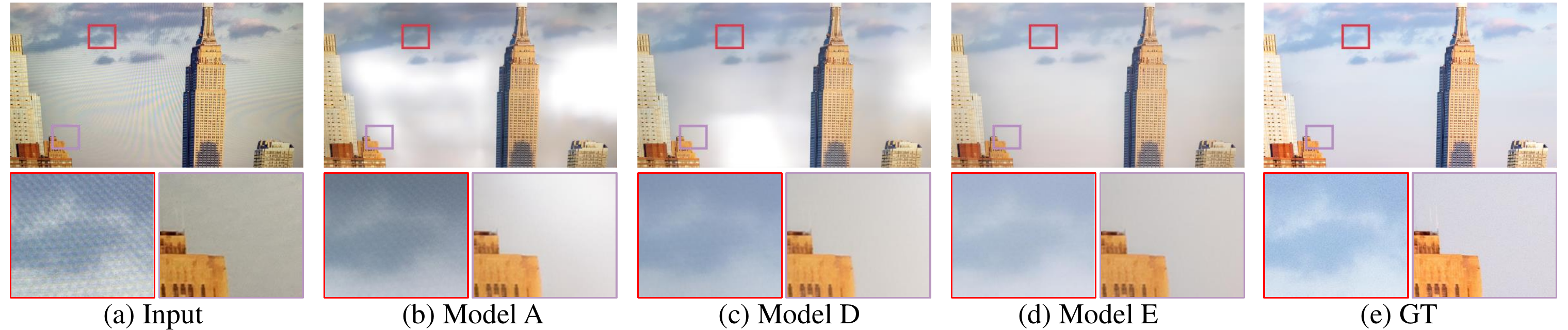}
    \caption{Qualitative effects of different components in SAM}
    \label{fig:ab} 
\end{figure} 

\subsection{Ablation Study}
In this section, we tease apart which components of our network contribute most to the final performance on the UHDM dataset. As shown in \cref{tab:ablation}, we start from the baseline model (model ``A''), which ablates the pyramid context extraction and the cross-scale dynamic fusion strategies. To make a fair comparison, we further build a stronger baseline model (model ``A$^+$'') that is comparable to our full model (model ``E'') in terms of the model capacity.

\noindent\textbf{Pyramid context extraction:}
We construct two variants (model ``B'' and model ``D'') for exploring the effectiveness of this design. Compared with the baseline (model ``A''), we observe that the proposed pyramid context extraction can significantly boost the model performance. To validate whether the improvement comes from more parameters in the additional two sub-branches, we exploit a weight-sharing strategy across all branches (model ``B''). The observations in \cref{tab:ablation} demonstrate that the performance gain mainly stems from the pyramid design rather than the increase of parameters. 
Further, as shown in \cref{fig:ab}, we find our pyramid design can successfully remove the moiré patterns that are not well addressed in the baseline model.

\noindent\textbf{Cross-scale dynamic fusion:}
To verify the importance of the proposed dynamic fusion scheme, we increasingly add this design to model ``B'' and model ``D'', resulting in model ``C'' and model ``E''. We observe consistent improvements for both models, especially on PSNR. Also, \cref{fig:ab} shows that the artifacts retained in model ``D'' are totally removed in the result of model ``E'', achieving a more harmonious color style.

\noindent\textbf{Loss function:} Through our experiments, we find the perceptual loss plays an essential role in image demoiréing. As shown in \cref{tab:ablation loss}, when replacing our loss function with a single $L_1$ loss, we notice obvious performance drops in our method, especially on LPIPS. Also, we make further exploration by applying our loss function to other state-of-the-art methods~\cite{sun2018moire,cheng2019multi}. The significant improvements on LPIPS illustrate the importance of the loss design in yielding a higher perceptual quality of recovered images. We suggest our loss is more robust to address the large-scale moiré patterns and the misaligned issue in the real-world datasets \cite{he2020fhde,sun2018moire}. More discussions are included in the supplementary file.

 

\begin{table*}[t]
    \caption{Ablation study of the loss function. The left and the right of ``/'' denote results trained by the pixel-wise $L_1$ loss and trained by our loss, respectively}
   \centering
   \renewcommand\tabcolsep{5.0pt}
   \resizebox{10cm}{!}{
   \begin{tabular}{c|c|ccc}
   \toprule[1.2pt]
   Dataset &Metrics   &DMCNN &MDDM  &Ours \\
   \hline \multirow{3}{*}{UHDM}
   &\text{PSNR}$\uparrow$ &\textbf{19.914}/19.911 &20.088/\textbf{20.333}   &21.489/\textbf{22.119}  \\
   
   &\text{SSIM}$\uparrow$ &\textbf{0.7575}/0.7212 & \textbf{0.7441}/0.7412  &0.7893/\textbf{0.7956}  \\
   
   &\text{LPIPS}$\downarrow$ &0.3764/\textbf{0.3089} &0.3409/\textbf{0.2986} &0.3330/\textbf{0.2551}  \\
\bottomrule[1.2pt]
  
\end{tabular}
}

\label{tab:ablation loss}
\end{table*}

\section{Conclusion}
In this paper, to explore the more practical yet challenging 4K image demoiréing scenario, we propose the first real-world ultra-high-definition demoiréing dataset (UHDM). Based upon this dataset, we conduct a benchmark study and limitation analysis of current methods, which motivates us to build a lightweight semantic-aligned scale-aware module (SAM) to strengthen the model's multi-scale ability without incurring much computational cost.
By leveraging SAM in different depths of a simple encoder-decoder backbone network, we develop ESDNet to handle 4K high-resolution image demoiréing effectively.
Our method is computationally efficient and easy to implement, achieving state-of-the-art results on four benchmark demoiréing datasets (including our UHDM). We hope our investigation could inspire future research in this more practical setting.

\subsubsection{Acknowledgements.} 
This work is partially supported by HKU-TCL Joint Research Center for Artificial Intelligence, Hong Kong Research Grant Council - Early Career Scheme (Grant No. 27209621), National Key R$\&$D Program of China (No.2021YFA1001300), and Guangdong-Hong Kong-Macau Applied Math Center grant 2020B1515310011.

\clearpage
%
%
\bibliographystyle{splncs04}
\bibliography{egbib}

\begin{thebibliography}{10}
\providecommand{\url}[1]{\texttt{#1}}
\providecommand{\urlprefix}{URL }
\providecommand{\doi}[1]{https://doi.org/#1}

\bibitem{anwar2020densely}
Anwar, S., Barnes, N.: Densely residual laplacian super-resolution. IEEE
  Transactions on Pattern Analysis and Machine Intelligence  (2020)

\bibitem{arjovsky2017wasserstein}
Arjovsky, M., Chintala, S., Bottou, L.: Wasserstein generative adversarial
  networks. In: International conference on machine learning. pp. 214--223.
  PMLR (2017)

\bibitem{cai2016unified}
Cai, Z., Fan, Q., Feris, R.S., Vasconcelos, N.: A unified multi-scale deep
  convolutional neural network for fast object detection. In: European
  conference on computer vision. pp. 354--370. Springer (2016)

\bibitem{chen2018learning}
Chen, C., Chen, Q., Xu, J., Koltun, V.: Learning to see in the dark. In:
  Proceedings of the IEEE Conference on Computer Vision and Pattern
  Recognition. pp. 3291--3300 (2018)

\bibitem{chen2017deeplab}
Chen, L.C., Papandreou, G., Kokkinos, I., Murphy, K., Yuille, A.L.: Deeplab:
  Semantic image segmentation with deep convolutional nets, atrous convolution,
  and fully connected crfs. IEEE transactions on pattern analysis and machine
  intelligence  \textbf{40}(4),  834--848 (2017)

\bibitem{chen2017photographic}
Chen, Q., Koltun, V.: Photographic image synthesis with cascaded refinement
  networks. In: Proceedings of the IEEE international conference on computer
  vision. pp. 1511--1520 (2017)

\bibitem{chen2018cascaded}
Chen, Y., Wang, Z., Peng, Y., Zhang, Z., Yu, G., Sun, J.: Cascaded pyramid
  network for multi-person pose estimation. In: Proceedings of the IEEE
  conference on computer vision and pattern recognition. pp. 7103--7112 (2018)

\bibitem{cheng2019multi}
Cheng, X., Fu, Z., Yang, J.: Multi-scale dynamic feature encoding network for
  image demoir{\'e}ing. In: 2019 IEEE/CVF International Conference on Computer
  Vision Workshop (ICCVW). pp. 3486--3493. IEEE (2019)

\bibitem{gao2019dynamic}
Gao, H., Tao, X., Shen, X., Jia, J.: Dynamic scene deblurring with parameter
  selective sharing and nested skip connections. In: Proceedings of the
  IEEE/CVF Conference on Computer Vision and Pattern Recognition. pp.
  3848--3856 (2019)

\bibitem{goodfellow2014generative}
Goodfellow, I., Pouget-Abadie, J., Mirza, M., Xu, B., Warde-Farley, D., Ozair,
  S., Courville, A., Bengio, Y.: Generative adversarial nets. Advances in
  neural information processing systems  \textbf{27} (2014)

\bibitem{gulrajani2017improved}
Gulrajani, I., Ahmed, F., Arjovsky, M., Dumoulin, V., Courville, A.: Improved
  training of wasserstein gans. arXiv preprint arXiv:1704.00028  (2017)

\bibitem{he2019mop}
He, B., Wang, C., Shi, B., Duan, L.Y.: Mop moire patterns using mopnet. In:
  Proceedings of the IEEE/CVF International Conference on Computer Vision. pp.
  2424--2432 (2019)

\bibitem{he2020fhde}
He, B., Wang, C., Shi, B., Duan, L.Y.: Fhde 2 net: Full high definition
  demoireing network. In: Computer Vision--ECCV 2020: 16th European Conference,
  Glasgow, UK, August 23--28, 2020, Proceedings, Part XXII 16. pp. 713--729.
  Springer (2020)

\bibitem{he2016deep}
He, K., Zhang, X., Ren, S., Sun, J.: Deep residual learning for image
  recognition. In: Proceedings of the IEEE conference on computer vision and
  pattern recognition. pp. 770--778 (2016)

\bibitem{huang2017densely}
Huang, G., Liu, Z., Van Der~Maaten, L., Weinberger, K.Q.: Densely connected
  convolutional networks. In: Proceedings of the IEEE conference on computer
  vision and pattern recognition. pp. 4700--4708 (2017)

\bibitem{johnson2016perceptual}
Johnson, J., Alahi, A., Fei-Fei, L.: Perceptual losses for real-time style
  transfer and super-resolution. In: European conference on computer vision.
  pp. 694--711. Springer (2016)

\bibitem{kim2016accurate}
Kim, J., Lee, J.K., Lee, K.M.: Accurate image super-resolution using very deep
  convolutional networks. In: Proceedings of the IEEE conference on computer
  vision and pattern recognition. pp. 1646--1654 (2016)

\bibitem{kingma2014adam}
Kingma, D.P., Ba, J.: Adam: A method for stochastic optimization. arXiv
  preprint arXiv:1412.6980  (2014)

\bibitem{lim2017enhanced}
Lim, B., Son, S., Kim, H., Nah, S., Mu~Lee, K.: Enhanced deep residual networks
  for single image super-resolution. In: Proceedings of the IEEE conference on
  computer vision and pattern recognition workshops. pp. 136--144 (2017)

\bibitem{liu2018demoir}
Liu, B., Shu, X., Wu, X.: Demoir$\backslash$'eing of camera-captured screen
  images using deep convolutional neural network. arXiv preprint
  arXiv:1804.03809  (2018)

\bibitem{liu2018image}
Liu, G., Reda, F.A., Shih, K.J., Wang, T.C., Tao, A., Catanzaro, B.: Image
  inpainting for irregular holes using partial convolutions. In: Proceedings of
  the European Conference on Computer Vision (ECCV). pp. 85--100 (2018)

\bibitem{liu2020wavelet}
Liu, L., Liu, J., Yuan, S., Slabaugh, G., Leonardis, A., Zhou, W., Tian, Q.:
  Wavelet-based dual-branch network for image demoir{\'e}ing. In: Computer
  Vision--ECCV 2020: 16th European Conference, Glasgow, UK, August 23--28,
  2020, Proceedings, Part XIII 16. pp. 86--102. Springer (2020)

\bibitem{loshchilov2016sgdr}
Loshchilov, I., Hutter, F.: Sgdr: Stochastic gradient descent with warm
  restarts. arXiv preprint arXiv:1608.03983  (2016)

\bibitem{pohlen2017full}
Pohlen, T., Hermans, A., Mathias, M., Leibe, B.: Full-resolution residual
  networks for semantic segmentation in street scenes. In: Proceedings of the
  IEEE conference on computer vision and pattern recognition. pp. 4151--4160
  (2017)

\bibitem{ronneberger2015u}
Ronneberger, O., Fischer, P., Brox, T.: U-net: Convolutional networks for
  biomedical image segmentation. In: International Conference on Medical image
  computing and computer-assisted intervention. pp. 234--241. Springer (2015)

\bibitem{shi2016real}
Shi, W., Caballero, J., Husz{\'a}r, F., Totz, J., Aitken, A.P., Bishop, R.,
  Rueckert, D., Wang, Z.: Real-time single image and video super-resolution
  using an efficient sub-pixel convolutional neural network. In: Proceedings of
  the IEEE conference on computer vision and pattern recognition. pp.
  1874--1883 (2016)

\bibitem{simonyan2014very}
Simonyan, K., Zisserman, A.: Very deep convolutional networks for large-scale
  image recognition. arXiv preprint arXiv:1409.1556  (2014)

\bibitem{song2018contextual}
Song, Y., Yang, C., Lin, Z., Liu, X., Huang, Q., Li, H., Kuo, C.C.J.:
  Contextual-based image inpainting: Infer, match, and translate. In:
  Proceedings of the European Conference on Computer Vision (ECCV). pp. 3--19
  (2018)

\bibitem{sun2018moire}
Sun, Y., Yu, Y., Wang, W.: Moir{\'e} photo restoration using multiresolution
  convolutional neural networks. IEEE Transactions on Image Processing
  \textbf{27}(8),  4160--4172 (2018)

\bibitem{suvorov2021resolution}
Suvorov, R., Logacheva, E., Mashikhin, A., Remizova, A., Ashukha, A.,
  Silvestrov, A., Kong, N., Goka, H., Park, K., Lempitsky, V.:
  Resolution-robust large mask inpainting with fourier convolutions. arXiv
  preprint arXiv:2109.07161  (2021)

\bibitem{tao2018scale}
Tao, X., Gao, H., Shen, X., Wang, J., Jia, J.: Scale-recurrent network for deep
  image deblurring. In: Proceedings of the IEEE Conference on Computer Vision
  and Pattern Recognition. pp. 8174--8182 (2018)

\bibitem{vedaldi2010vlfeat}
Vedaldi, A., Fulkerson, B.: Vlfeat: An open and portable library of computer
  vision algorithms. In: Proceedings of the 18th ACM international conference
  on Multimedia. pp. 1469--1472 (2010)

\bibitem{wang2020deep}
Wang, J., Sun, K., Cheng, T., Jiang, B., Deng, C., Zhao, Y., Liu, D., Mu, Y.,
  Tan, M., Wang, X., et~al.: Deep high-resolution representation learning for
  visual recognition. IEEE transactions on pattern analysis and machine
  intelligence  \textbf{43}(10),  3349--3364 (2020)

\bibitem{wang2020vcnet}
Wang, Y., Chen, Y.C., Tao, X., Jia, J.: Vcnet: A robust approach to blind image
  inpainting. In: Computer Vision--ECCV 2020: 16th European Conference,
  Glasgow, UK, August 23--28, 2020, Proceedings, Part XXV 16. pp. 752--768.
  Springer (2020)

\bibitem{wang2004image}
Wang, Z., Bovik, A.C., Sheikh, H.R., Simoncelli, E.P.: Image quality
  assessment: from error visibility to structural similarity. IEEE transactions
  on image processing  \textbf{13}(4),  600--612 (2004)

\bibitem{xie2019image}
Xie, C., Liu, S., Li, C., Cheng, M.M., Zuo, W., Liu, X., Wen, S., Ding, E.:
  Image inpainting with learnable bidirectional attention maps. In: Proceedings
  of the IEEE/CVF International Conference on Computer Vision. pp. 8858--8867
  (2019)

\bibitem{yang2017high}
Yang, C., Lu, X., Lin, Z., Shechtman, E., Wang, O., Li, H.: High-resolution
  image inpainting using multi-scale neural patch synthesis. In: Proceedings of
  the IEEE conference on computer vision and pattern recognition. pp.
  6721--6729 (2017)

\bibitem{yeh2016semantic}
Yeh, R., Chen, C., Lim, T.Y., Hasegawa-Johnson, M., Do, M.N.: Semantic image
  inpainting with perceptual and contextual losses. arXiv preprint
  arXiv:1607.07539  \textbf{2}(3) (2016)

\bibitem{yu2015multi}
Yu, F., Koltun, V.: Multi-scale context aggregation by dilated convolutions.
  arXiv preprint arXiv:1511.07122  (2015)

\bibitem{yuan2019aim}
Yuan, S., Timofte, R., Slabaugh, G., Leonardis, A., Zheng, B., Ye, X., Tian,
  X., Chen, Y., Cheng, X., Fu, Z., et~al.: Aim 2019 challenge on image
  demoireing: Methods and results. In: 2019 IEEE/CVF International Conference
  on Computer Vision Workshop (ICCVW). pp. 3534--3545. IEEE (2019)

\bibitem{zamir2021multi}
Zamir, S.W., Arora, A., Khan, S., Hayat, M., Khan, F.S., Yang, M.H., Shao, L.:
  Multi-stage progressive image restoration. In: Proceedings of the IEEE/CVF
  Conference on Computer Vision and Pattern Recognition. pp. 14821--14831
  (2021)

\bibitem{zhang2019deep}
Zhang, H., Dai, Y., Li, H., Koniusz, P.: Deep stacked hierarchical multi-patch
  network for image deblurring. In: Proceedings of the IEEE/CVF Conference on
  Computer Vision and Pattern Recognition. pp. 5978--5986 (2019)

\bibitem{zhang2017beyond}
Zhang, K., Zuo, W., Chen, Y., Meng, D., Zhang, L.: Beyond a gaussian denoiser:
  Residual learning of deep cnn for image denoising. IEEE transactions on image
  processing  \textbf{26}(7),  3142--3155 (2017)

\bibitem{zhang2018unreasonable}
Zhang, R., Isola, P., Efros, A.A., Shechtman, E., Wang, O.: The unreasonable
  effectiveness of deep features as a perceptual metric. In: Proceedings of the
  IEEE conference on computer vision and pattern recognition. pp. 586--595
  (2018)

\bibitem{zhang2019zoom}
Zhang, X., Chen, Q., Ng, R., Koltun, V.: Zoom to learn, learn to zoom. In:
  Proceedings of the IEEE/CVF Conference on Computer Vision and Pattern
  Recognition. pp. 3762--3770 (2019)

\bibitem{zhang2018residual}
Zhang, Y., Tian, Y., Kong, Y., Zhong, B., Fu, Y.: Residual dense network for
  image super-resolution. In: Proceedings of the IEEE conference on computer
  vision and pattern recognition. pp. 2472--2481 (2018)

\bibitem{zheng2020image}
Zheng, B., Yuan, S., Slabaugh, G., Leonardis, A.: Image demoireing with
  learnable bandpass filters. In: Proceedings of the IEEE/CVF Conference on
  Computer Vision and Pattern Recognition. pp. 3636--3645 (2020)

\bibitem{zhou2018stereo}
Zhou, T., Tucker, R., Flynn, J., Fyffe, G., Snavely, N.: Stereo magnification:
  Learning view synthesis using multiplane images. arXiv preprint
  arXiv:1805.09817  (2018)

\end{thebibliography}

\title{Supplementary Material for\\
 ``Towards Efficient and Scale-Robust Ultra-High-Definition Image Demoiréing''} 
\titlerunning{Supp. File of Towards Efficient and Scale-Robust UHD Image Demoiréing}
\renewcommand{\thefootnote}{\textrm{\Letter}} 

\author{}
\institute{}
\authorrunning{X. Yu et al.}

\maketitle
   
\renewcommand*{\thesection}{\Alph{section}}
\setcounter{section}{0}

\section*{Outline}
We supplement the main body of our paper with additional details, discussions, and results in this document.
In~\cref{sec:supp-dataset}, we present more details about our dataset capture, which includes a brief analysis of the formation of degraded screen images.
In~\cref{sec:supp-method}, we provide more implementation details of our network architecture as well as a simple empirical study of loss functions to assist us in selecting a suitable training objective for moiré removal.
In~\cref{sec:result}, we provide more implementation details of experiments and show more qualitative results and comparisons with other state-of-the-art methods.
Furthermore, as shown in~\cref{sec:fhd}, we investigate why FHDe$^2$Net fails on this more challenging 4K dataset.  
We conduct a more detailed discussion of current methods' strategies for handling scale-variation of moiré patterns in~\cref{sec:revisit}.

\section{Dataset Capture and Analysis}\label{sec:supp-dataset}
In this section, we first present a brief introduction of the formation of the moiré pattern, and then we provide more details about our capture settings.

\subsection{Image Degradation Analysis}
The formation of degraded screen images taken with mobile devices can be divided into two processes: the generation of moiré patterns caused by frequency aliasing; and the global color degradation of the image, caused by a series of ISP operations (\eg, auto exposure control, white balance correction, gamma correction, and global tone mapping).

\begin{figure}[t]
\begin{minipage}[t]{0.5\textwidth}
\centering
\includegraphics[width=5cm]{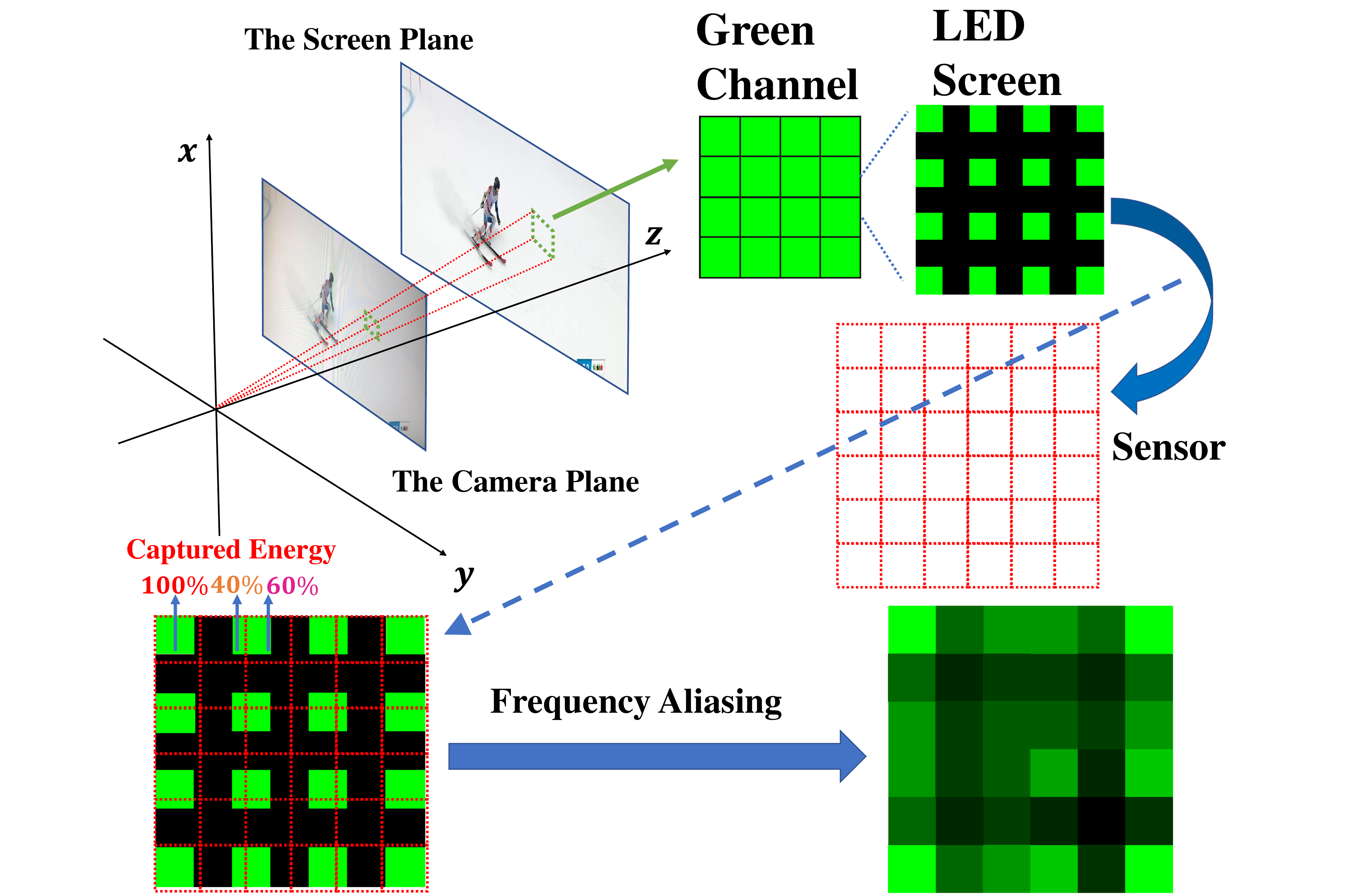}
\end{minipage}
\begin{minipage}[t]{0.5\textwidth}
\centering
\includegraphics[width=6cm]{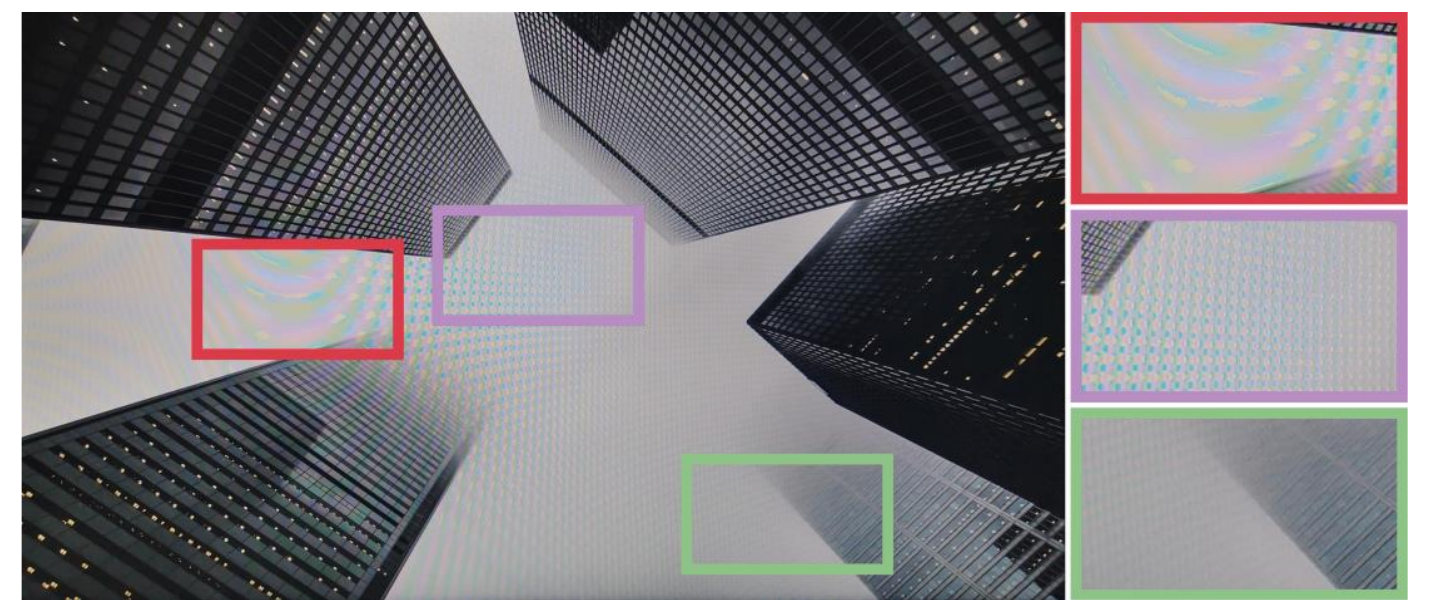}
\end{minipage}
\caption{Left: the formation of the moiré pattern.  Notice that there are small gaps between the light-emitting diodes. Right: the characteristics of moiré patterns}
\label{fig:principle}

\end{figure}

We can model the formation of moiré patterns as a local color unbalanced scaling in the camera's color filter array (CFA). 
Without loss of generality, consider how one of the green channels in the RGBG raw pattern is collected. 
As shown in \cref{fig:principle}, due to a slight misalignment between sensor pixels and LED screen pixels, the energy may shift from one pixel to its neighbors. This flow eventually aligns again after passing a few pixels. Hence, the value of each pixel in this period could be modeled as being multiplied by different scaling factors:
\begin{equation}
\hat{R}(i, j)=R(i, j) * S(i, j), \\
\end{equation}
where $\hat{R}(i, j)=\left(\hat{r}_{ij}, \hat{g}_{ij}^{1}, \hat{b}_{ij}, \hat{g}_{ij}^{2}\right)$ represents the degraded pixel at the location $\left(i,j\right)$ in the Bayer pattern and
$R(i, j)$ denotes the clean pixel. $S(i,j)=\left(s_{ij}^{r}, s_{ij}^{g_{1}}, s_{ij}^{b}, s_{ij}^{g_{2}}\right)$ is the scaling factor for four channels (RGBG) caused by frequency aliasing; $*$ denotes the point-wise multiplication. However, since the LED display and camera Bayer array both emit or receive each channel information in an alternate form, the scaling rules for different channels are not consistent within a cycle. Hence, the camera stores a wrong color distribution, causing the moiré pattern we see. 

Furthermore, there is an unavoidable gap when we re-capture an image on the screen. For instance, ambient light can lead to incorrect exposure control, wrong auto white balance, and unnatural tone mapping. Also, corrupted raw data can affect the process of raw image demosaicing. All of these factors contribute to the overall degradation of the color, which can be formulated as:
\begin{equation}
M=F\left(R * S\right), \\
\end{equation}
where $M$ is the final degraded screen image and $F$ is a nonlinear function that globally affects the image quality.

Given the above analysis, we could explain the following characteristics (see \cref{fig:principle}) of moiré patterns:

\noindent\textbf{Structural distortions:}
Since the RGB color distributions change in an alternate form, the local illuminance contrasts among the three channels are not consistent. Thus, new structures are created and mixed with original contents.

\noindent\textbf{Diverse degraded forms:}
In \cref{fig:principle}, we show the simplest case of misalignment between two patterns, in which the camera plane and the screen plane are parallel to each other. Obviously, the scaling rule would be quite different if the angle and distance between these two planes were to change, resulting in moiré patterns in different shapes and scales. This explains why the moiré pattern characteristics highly depend on the geometric relationship between the screen and the camera.

\noindent\textbf{Large-scale patterns in low-frequency regions:}
Unlike the natural image captured from real scenes, we capture discrete signals emitted from the LED screen and store them in new discrete forms. Thus, the low-frequency image areas actually become signals with the highest frequency and are more likely to continuously alias with the camera sensor over a long period, resulting in larger moiré patterns.

\subsection{More Details about Capture Settings}
\begin{figure}
   \begin{center}
      \includegraphics[width=1\textwidth]{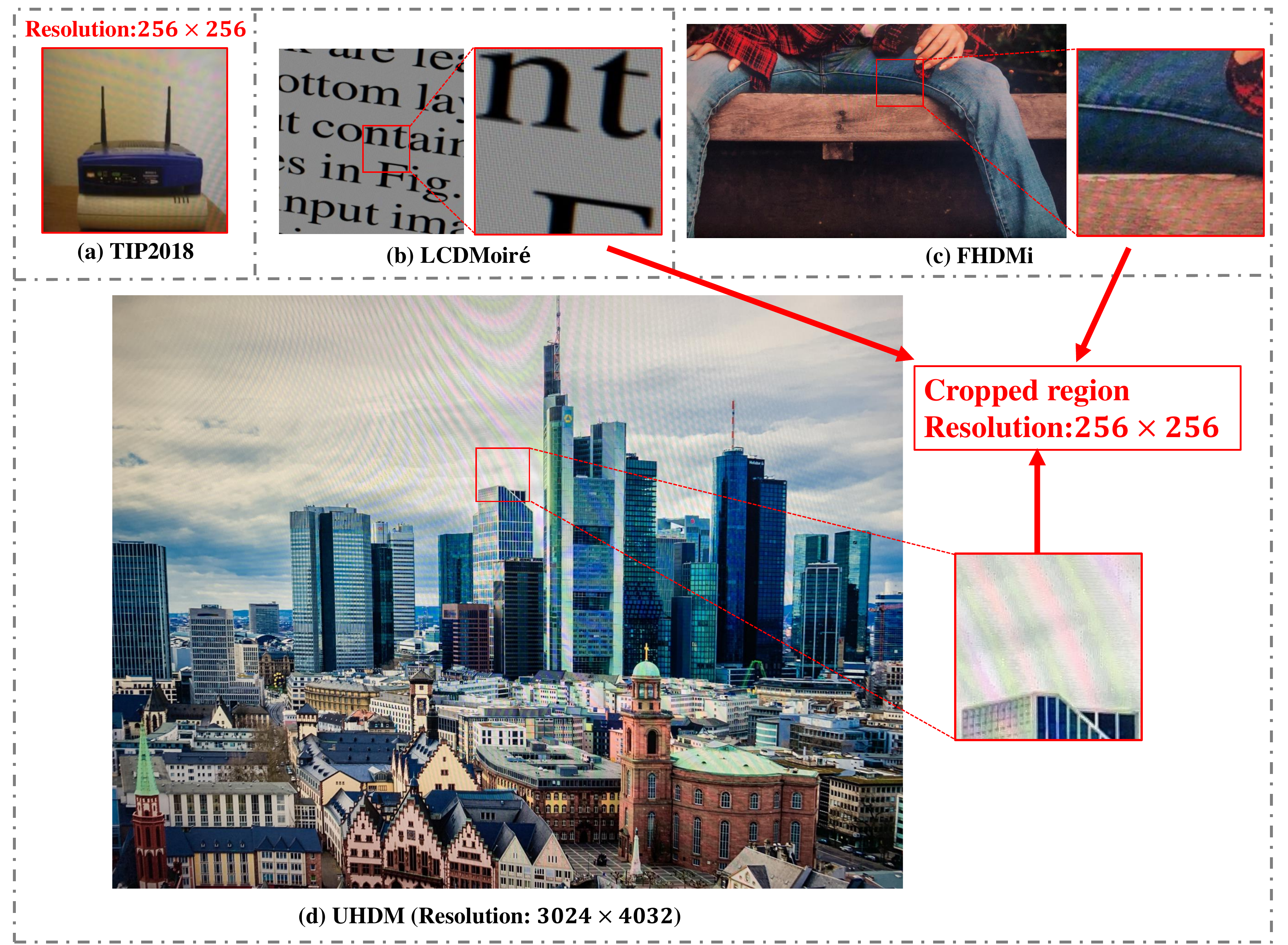}
   \end{center}
      
      \caption{Comparisons with other datasets; we crop patches from these four datasets at the same resolution $256\times 256$ (the image in TIP2018 dataset~\cite{sun2018moire} is already at a resolution of $256\times 256$)}
      
   \label{fig:data_compare}
\end{figure}

Based on the above analysis, we thus shoot the screen images via different camera views to produce different patterns and combine multiple devices to produce diverse degradation styles (including pattern appearance and global color style). 
Specifically, we apply three mobile phones and three digital screens, as shown in \cref{tab:devices} ($3\times 3=9$ combinations here totally). 
Notably, the ``4K'' challenge means the obtained moiré image is at a resolution of ultra-high-definition (\ie, the shooting resolution is 4K). We also compare our dataset with other datasets visually. As seen in \cref{fig:data_compare}, we crop patches from these four datasets at the same resolution $256\times 256$ (the image in TIP2018 dataset~\cite{sun2018moire} is already at a resolution of $256\times 256$). Obviously, compared with other datasets, the image UHDM suffers from more severe moiré artifacts and has less clean image content to harvest in a local window. As a result, it is more challenging for the network to identify the moiré pattern or fill clean content into the degraded region, which has also been demonstrated in~\cite{he2020fhde}.    

\begin{table}
\caption{The capture devices we apply to get the moiré image}
\centering
\resizebox{9cm}{!}{
\begin{tabular}{c|c|c|c}
\toprule[1pt]
Mobile Phone  &Shooting Resolution &Digital Screen  &Display Resolution \\
\hline 
iPhone XR&$4032\times 3024$ &LG 27UL650-W &$3840\times 2160$   \\

iPhone 13&$4032\times 3024$ &AOC U2790PQU &$3840\times 2160$   \\

Redmi K30 Pro&$4624\times 3472$ &Philips 243S7EHMB &$1920\times 1080$  \\

\bottomrule[1pt]
\end{tabular}
}

\label{tab:devices}
\end{table}

\section{Method}\label{sec:supp-method}
\begin{figure}
      \includegraphics[width=1\textwidth]{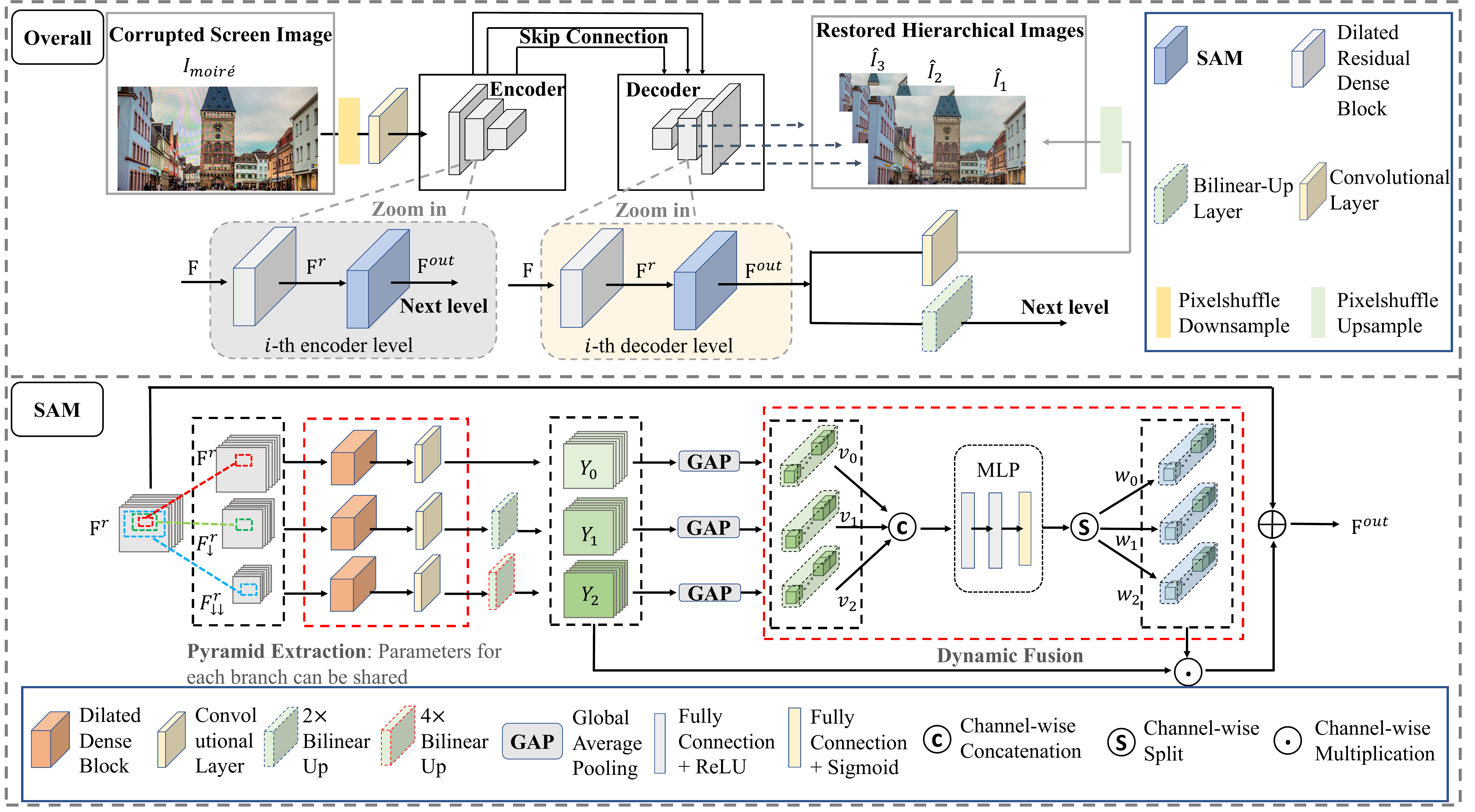}
      \caption{The pipeline of our ESDNet and the proposed semantic-aligned scale-aware module (SAM) 
      }
\label{fig:network}
\end{figure}

In this section, we give details of our network architecture. The overview of our network is shown in \cref{fig:network}. We use skip-connections to connect each level of the encoder and decoder, wherein the features are concatenated. 

\subsection{Semantic-Aligned Scale-Aware Module (SAM)}
As seen in \cref{fig:network}, there are three branches in the pyramid context extraction module wherein the dilated dense block ($L=5$) is utilized as the backbone block to extract the context information. Two bilinear upsampling layers with upsampling ratios $2$ and $4$ are applied to the second and third branches to align the spatial resolution of the first branch. There are three fully connected layers for the MLP in the cross-scale dynamic fusion module to learn the adaptive weights. We adopt ReLU for the first two layers and Sigmoid for the last layer as our nonlinear activation functions. Specifically, for an input tensor $v\in\mathbb{R}^{{1}\times{1}\times{3C}}$, the channel number is squeezed by a dividing factor $4$ in the first layer and then expanded to the original number in the last layer.

\noindent\textbf{Weight-sharing SAM:}
We apply a weight-sharing strategy for one of our models, denoted as WS-ESDNet, which shares the learnable parameters among the three branches. The WS-ESDNet has fewer parameters while keeping comparable quantitative and qualitative results compared to our standard model ESDNet. The quantitative results have already been shown in Section.\textcolor{red}{5.2} in the main body of our paper, and qualitative results are illustrated in \cref{sec:result}. This demonstrates that the performance gain primarily benefits from our architecture design rather than increased model parameters.

\begin{table*}
\centering
\caption{Detail of the encoder; DRDB denotes the dilated residual dense block consisting of three convolution layers}
\renewcommand\tabcolsep{5.0pt}
\resizebox{12cm}{!}{
\begin{tabular}{|c|c|cccc|}
\hline Level & Block Type & Input Channels & Output Channels&Inter Channels &Dilation Rates \\
\hline 1 & Pixelshuffle downsampling & 3  & 12& - & - \\
& $5\times 5$ Conv + ReLU & 12  & 48 & - & -\\
& DRDB & 48  & 48 & 32 & (1, 2, 1)\\
& SAM & 48  & 48 & 32 & (1, 2, 3, 2, 1)\\
\hline 2 & Stride=2, $3\times 3$ Conv & 48  & 96 & - & -\\
& DRDB & 96  & 96& 32 & (1, 2, 1) \\
& SAM & 96 & 96 & 32 & (1, 2, 3, 2, 1)\\
\hline 3 & Stride=2, $3\times 3$ Conv & 96  & 192 & - & -\\
& DRDB & 192  & 192 & 32 & (1, 2, 1)\\
& SAM & 192 & 192 & 32 & (1, 2, 3, 2, 1)\\
\hline
\end{tabular}
}
\label{tab:encoder}
\end{table*}

\begin{table*}
\centering
\caption{Detail of the decoder; DRDB denotes the dilated residual dense block consisting of three convolution layers}
\renewcommand\tabcolsep{5.0pt}
\resizebox{12cm}{!}{
\begin{tabular}{|c|c|cccc|}
\hline Level & Block Type & Input Channels & Output Channels &Inter Channels &Dilation Rates\\

\hline 3 & $3\times 3$ Conv + ReLU & 192  & 64 & - & -\\
& DRDB & 64  & 64  & 32 & (1, 2, 1)\\
& SAM & 64 & 64 & 32 & (1, 2, 3, 2, 1)\\
\hdashline 
Output Layer & $3\times 3$ Conv & 64 & 12 & -& -\\
& Pixelshuffle upsampling & 12 & 3& -& - \\
\hdashline
Transition Layer& Bilinear-Up Layer & 64 & 64& -& - \\

\hline 2 & $3\times 3$ Conv + ReLU & 160  & 64& -& - \\
& DRDB & 64  & 64 & 32& (1, 2, 1)\\
& SAM & 64 & 64 & 32& (1, 2, 3, 2, 1)\\
\hdashline 
Output Layer & $3\times 3$ Conv & 64 & 12& -& - \\
& Pixelshuffle upsampling & 12 & 3& -& - \\
\hdashline
Transition Layer& Bilinear-Up Layer & 64 & 64 & -& -\\

\hline 1 & $3\times 3$ Conv + ReLU & 112  & 64& -& - \\
& DRDB & 64  & 64 & 32& (1, 2, 1)\\
& SAM & 64 & 64 & 32& (1, 2, 3, 2, 1)\\
\hdashline 
Output Layer & $3\times 3$ Conv & 64 & 12 & -& -\\
& Pixelshuffle upsampling & 12 & 3 & -& -\\
\hdashline
Transition Layer& Bilinear-Up Layer & 64 & 64 & -& -\\

\hline
\end{tabular}
}
\label{tab:decoder} 
\end{table*}

\subsection{Empirical Study of Loss Functions}\label{sec:loss}

The loss function plays an essential role in guiding model updates and encouraging the model to learn natural patterns from data. To this end, we carry out an empirical study to investigate the impacts of different loss functions on image demoiréing. 

\begin{figure}
   \begin{center}
      \includegraphics[width=0.7\textwidth]{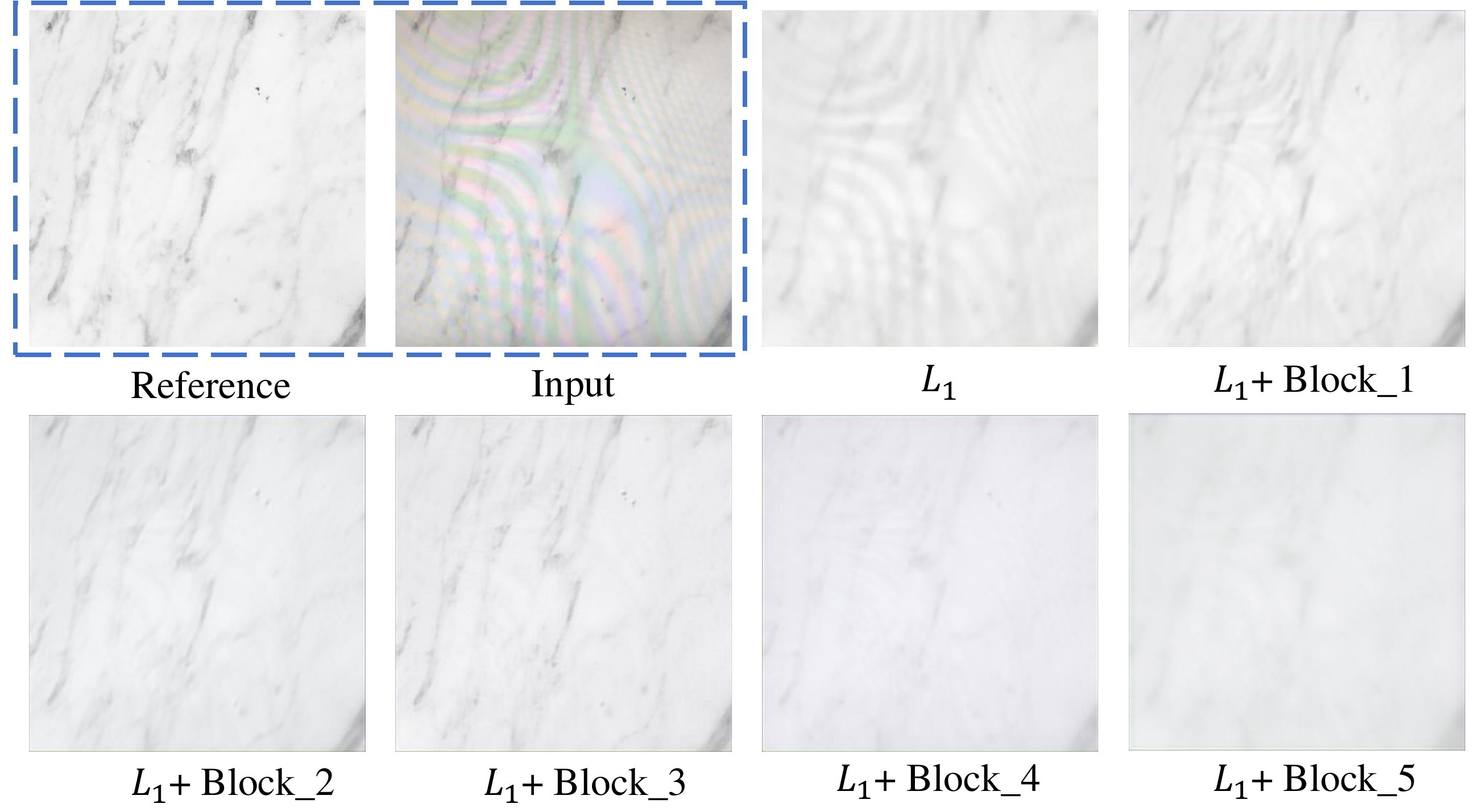}
   \end{center}
      
      \caption{The optimal results by fitting different loss functions for a single moiré image}
      
   \label{fig:loss study}
\end{figure}

We evaluate traditional $L_1$ loss and its combination with perceptual losses~\cite{johnson2016perceptual} where the features are respectively from the end of block\_1, block\_2, block\_3, block\_4 and block\_5 of a pre-trained VGG-16 network~\cite{simonyan2014very}.
We develop a simple task to study the effectiveness of these loss functions on removing undesirable moiré patterns. 
Specifically, we choose a degraded screen image $M$ with severe structural distortions and its corresponding clean ground-truth $I$; our aim is to restore $M$ by  optimizing $\theta^{\star}=\arg \min _{\theta} D(I, f_{\theta}(M))$ through our designed network $f_{\theta}$, where $D$ denotes the loss function, and $\hat{I} = f_{\theta^{\star}}(M)$ is the recovered image.
As shown in \cref{fig:loss study}, the single $L_1$ loss or its combination with the shallow block\_1 perceptual loss cannot guide the network to remove unnecessary structures; they are effective in restoring the pixel-level color due to their low-level nature. 
Meanwhile, the loss functions derived from block\_4 and block\_5 features, containing too deep semantic-level information, will lead the predicted image to lose its textures. 
In contrast, perceptual loss with features from  block\_2 and block\_3 can encourage the network to remove undesirable structures while preserving the original texture, a good signal for image demoiréing.  
In particular, the model trained with  block\_3 recovers more details with satisfying local contrasts. Hence, the block\_3 might be the most suitable layer to construct the training objective.   

Although many previous works~\cite{he2019mop,he2020fhde,liu2020wavelet} have already adopted the perceptual loss as a regularization term, they often overlook the importance of precisely choosing a suitable layer for this specific task, which is crucial, as different features will encourage the network to optimize the network in different directions.

\section{Experiments}\label{sec:result}
\subsection{Implementation Details}
We implement all the experiments using PyTorch on an NVIDIA RTX 3090 GPU card. The learning rate is initially set to $0.0002$ and scheduled by cyclic cosine annealing \cite{loshchilov2016sgdr}, and models are optimized by Adam \cite{kingma2014adam} with $\beta_{1}= 0.9 $ and $\beta_{2} = 0.999$. For UHDM dataset, we set the batch size as $2$. Notably, we conduct benchmark implementations of other methods~\cite{sun2018moire,zheng2020image,he2019mop,he2020fhde,cheng2019multi,liu2020wavelet} on our dataset sufficiently. For DMCNN~\cite{sun2018moire}, MDDM~\cite{cheng2019multi}, WDNet~\cite{liu2020wavelet} and MBCNN~\cite{zheng2020image}, we randomly crop a $768 \times 768$ patch from the ultra-high-definition images, and train the model for $150$ epochs, i.e., the totally same setting with ours. For FHDe$^2$Net~\cite{he2020fhde}, due to its different multi-stage nature and high computational cost, we can only follow its default setting in the official released code for training (\ie, 
down-sampled-resolution $384\times 384$ for training its global stage and cropped $384\times 384$ region for training the following three cascaded networks). For MopNet\cite{he2019mop}, we freeze its pre-trained classification sub-network and train its edge-prediction sub-network and demoiréing sub-network for 150 epochs, wherein we also crop a $384\times 384$ region for training. During inference, since MopNet cannot directly process the 4K image due to its heavy memory cost, we downsample the input image into 1080p (the highest resolution it can process on a single GPU) resolution and then upsample the result back to 4K resolution.   

\noindent\textbf{Other datasets:}
For FHDMi~\cite{he2020fhde} and LCDmoiré~\cite{yuan2019aim} dataset, we randomly crop a $512 \times 512$ patch from the high-definition images, and train the model for $150$ epochs with the batch size as 2. For TIP2018 dataset~\cite{sun2018moire}, we follow the benchmark setting, \ie, we first resize the image into a $286\times 286$ resolution and then do center crop to produce a $256\times 256$ resolution image for both training and testing. We train our models for 70 epochs and set batch size to 4. 

\subsection{Discussion about FHDe$^2$Net}
\label{sec:fhd}

We find that in the new dataset UHDM, FHDe$^2$Net suffers from a more significant performance drop than other methods. To this end, we conduct a parameters searching and analysis. Specifically, since we find the key challenge is to fuse the high-frequency detail, we mainly analyze the training of the last stage, \ie, the FDN and FRN (please refer to \cite{he2020fhde} for more details). Since the learning rate is scheduled by cyclic cosine annealing, which warms up every 50 epochs,  we evaluate the performances after the FDN and FRN (the last stage of FHDe$^2$Net) have been trained for 50, 100, and 150 epochs, respectively. As shown in \cref{tab:epoch}, with the increase of training time, SSIM improves significantly, but LPIPS degrades simultaneously. 
For this phenomenon, we attribute the reasons to two aspects, as elaborated upon below.

\begin{figure}
   \begin{center}
      \includegraphics[width=1\textwidth]{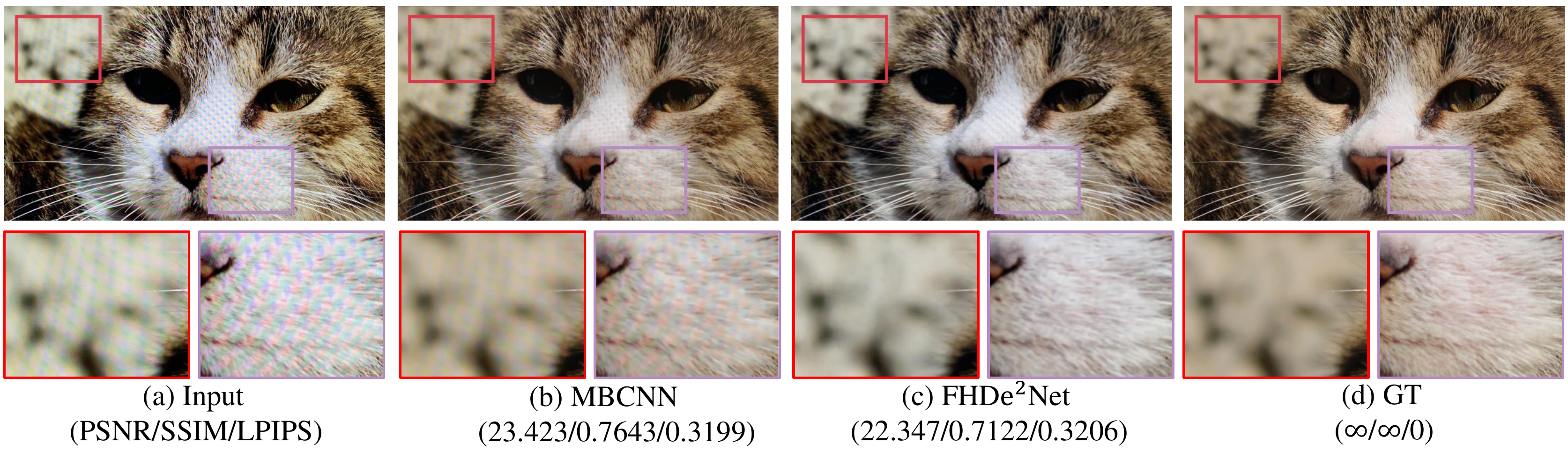}
   \end{center}
      
      \caption{Current metrics have some limitations. In this case, FHDe$^2$Net removes the moiré pattern more cleanly yet is still behind the MBCNN if evaluated by the three metrics }
      
   \label{fig:metric}
\end{figure}

On the one hand, current low-level metrics have several limitations and cannot fully measure the demoiréing performance (see \cref{fig:metric}). For example, PSNR is a pixel-wise metric sensitive to pixel misalignment and slight color shift, which has limited effect in measuring the structural distortion caused by the moiré pattern. SSIM is more robust to evaluate structural distortion yet still sensitive to the unstructured distortion (\eg,  pixel shift, rotation.), which is unavoidable in real-world data pairs. LPIPS has been proven to be more consistent with human perception; however, it is sensitive to blur as demonstrated in \cite{zhang2018unreasonable}. 

\begin{figure}
   \begin{center}
      \includegraphics[width=1\textwidth]{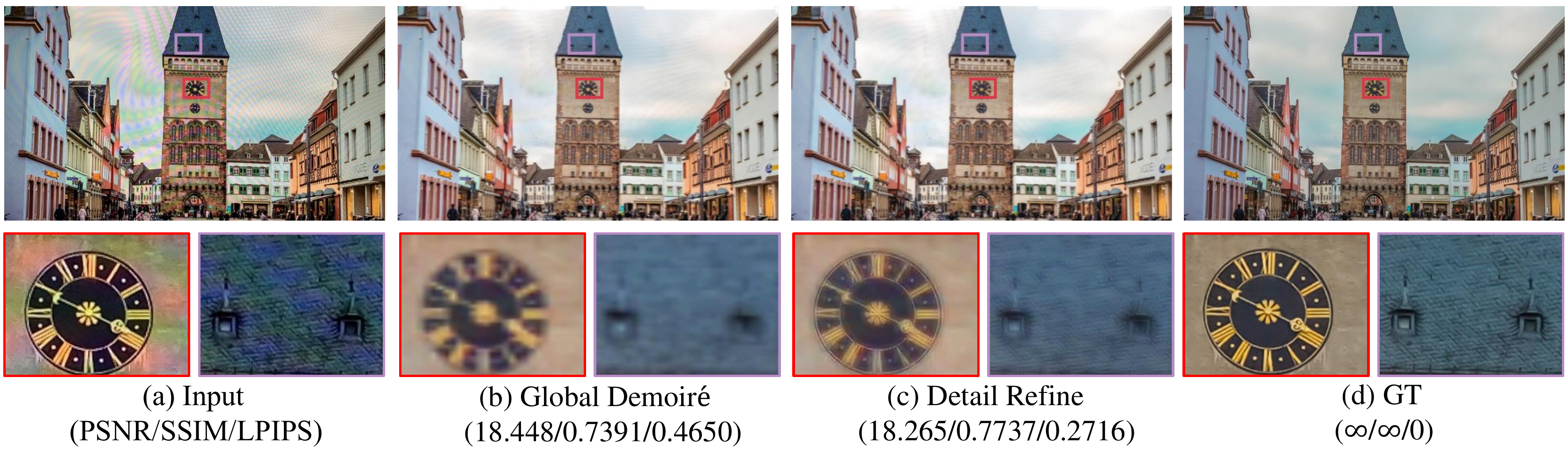}
   \end{center}
      
      \caption{Comparisons between the result produced by global demoiréing stage and the final result (\ie, ``Detail Refine''), in which the PSNR is almost unchanged while LPIPS achieves significant improvement }
      
   \label{fig:up}
\end{figure}

On the other hand, this indicates FHDe$^2$Net has reached its limit in making the trade-off between large-scale moiré removal and high-frequency details preservation. To explore whether this stage plays a role in high-frequency detail recovery, we compare it with the initial low-resolution result produced by the global demoiréing stage. As shown in \cref{tab:epoch}, compared with the initial result (\ie,  ``Low-resolution''), the fine-tuned model (\ie,  ``150 epoch'') achieves a significant improvement in LPIPS which indicates the detail has been recovered to some degree (but not been fully recovered, see \cref{fig:up}). However, the PSNR is almost unchanged, indicating that this stage may not work well for color recovery. 
One possible reason is that the fusion stage only utilizes the Y-channel's information of the original high-resolution image but lacks UV-channels' high-resolution information. Besides, to avoid the effect of pixel misalignment, FHDe$^2$Net does not adopt pixel-wise loss terms (\eg, $L_1, L_2$), which may prevent it from recovering the global color style.  
Under this circumstance, the accurate color information loses significantly, negatively affecting all three metrics, especially for the PSNR.

\begin{table}[H]
\caption{Quantitative results of different implementations of FHDe$^2$Net on UHDM dataset. ``Pre-train'' denotes the inference result by directly applying the official released pre-train model on FHDMi dataset~\cite{he2020fhde}, ``Low-resolution'' denotes the intermediate result produced by the first global demoiréing stage in FHDe$^2$Net}
\centering
\resizebox{9cm}{!}{
\begin{tabular}{c|cccccc}
\toprule[1.2pt]
Metrics  &Input &Pre-train &Low-resolution &50 epoch &100 epoch &150 epoch   \\
\hline 
\text{PSNR}$\uparrow$ &17.117 &18.052 &20.333 &20.312 &20.313 &20.338   \\

\text{SSIM}$\uparrow$ &0.5089 &0.5986 &0.7408 &0.7290 &0.7365 &0.7496   \\

\text{LPIPS}$\downarrow$ &0.5314 &0.4929 &0.4669 &0.3397 &0.3429 &0.3519  \\

\bottomrule[1.2pt]
\end{tabular}
}

\label{tab:epoch}
\end{table}

In fact, we have conducted several parameters searching for the last stage's training (consists of two sub-networks FDN and FRN), trying to improve the performance of FHDe$^2$Net. To be precise, we adjust the loss weights to guide the networks' optimization. As illustrated in Eq.\eqref{eq:last}, the overall loss function of the last stage consists of two parts: $L_{\text{FDN}}$ and $L_{\text{FRN}}$, where $L_{\text{FDN}}$ aims to reconstruct the high-resolution gray-scale image (\ie, the Y-channel of YUV color space) and $L_{\text{FRN}}$ aims to further fuse the color information (more details can be referred to \cite{he2020fhde}): 

\begin{equation}
\mathcal{L}_{\text{last}}(I, \hat{I}) = \mathcal{L}_{\text{FDN}}(I_{Y}, \hat{I}_{Y})+\lambda\times\mathcal{L}_{\text{FRN}}(I, \hat{I})
\label{eq:last}
\end{equation}
where $I$ is the ground-truth and $\hat{I}$ is the network's output, $I_Y$ and $\hat{I}_Y$ denote their Y-channel components, respectively. 
Moreover, for $L_{\text{FRN}}$, it is essentially a CoBi~\cite{zhang2019zoom} loss, which aims to measure the similarity between unaligned image pairs, consisting of a term $\mathbb{D}$ to measure feature similarity and a term $\mathbb{D}^{\prime}$ to compute the spatial distance between these two pixels (with a weight $w_s$), \ie,:

\begin{equation}
\mathcal{L}_{\text{FRN}}(\hat{I}, I)=\frac{1}{N} \sum_{i}^{N} \min _{j=1, \ldots, M}\left((1-w_s)\mathbb{D}\left(p_{i}, q_{j}\right)+w_{s} \mathbb{D}^{\prime}\left(p_{i}, q_{j}\right)\right)
\end{equation}
where $p_i, q_j$ stand for the feature vectors from the output image $\hat{I}$ and clean image $I$ at the spatial position indexed by $i$ and $j$, respectively. $N$, $M$ denote the amounts of features (\ie, the amounts in searching space).

We try several $(\lambda, w_s)$ combinations to train the model. For fast exploration, we train every model for 50 epochs and compare their results, as shown in \cref{tab:FHD}. However, since the metrics' changes of each model are not significant, we use the default parameter settings to report the results in our main paper.  

In summary, although FHDe$^2$Net achieves the best (except for ours) result on the FHDMi dataset~\cite{he2020fhde}, this framework is not robust under the higher resolution setting. Moreover, its complex module designs further render it hard to be applied to the 4K scenario due to unacceptable increased computational costs. 

\begin{table}[H]
\caption{Quantitative comparisons of different weights for training FHDe$^2$Net. ``A'' denotes the default model where $(\lambda, w_s)=(1, 0.5)$; ``B'' denotes $(\lambda, w_s)=(0.5, 0.5)$; ``C'' denotes $(\lambda, w_s)=(2, 0.5)$;  ``D'' denotes $(\lambda, w_s)=(1, 0.7)$; ``E'' denotes $(\lambda, w_s)=(1, 0.2)$}
\centering
\resizebox{10cm}{!}{
\begin{tabular}{c|ccccccc}
\toprule[1.2pt]
Metrics  &Input &Pre-train  &Model A &Model B &Model C  &Model D &Model E   \\
\hline 
\text{PSNR}$\uparrow$ &17.117 &18.052 &20.312 &20.282 &20.174 &20.251 &19.050 \\

\text{SSIM}$\uparrow$ &0.5089 &0.5986 &0.7290 &0.7392 &0.7350 &0.7435 &0.7240 \\

\text{LPIPS}$\downarrow$ &0.5314 &0.4929 &0.3397 &0.3409 &0.3359 &0.3497 &0.3566 \\

\bottomrule[1.2pt]
\end{tabular}
}

\label{tab:FHD}
\end{table}

\subsection{SAM for Other Methods}
We demonstrate that equipping with the proposed SAM can also help other methods to achieve performance gain. Here we conduct experiments on MDDM~\cite{cheng2019multi}, DMCNN~\cite{sun2018moire} and MBCNN~\cite{zheng2020image}, where we stack SAM in these networks. As shown in \cref{tab:SAM}, all metrics have improvements.   

\begin{table}[H]
\caption{Effects of the proposed SAM. We add our SAM to current methods DMCNN~\cite{sun2018moire}, MDDM~\cite{cheng2019multi} and MBCNN~\cite{zheng2020image} to improve their performances}
\centering
\resizebox{10cm}{!}{
\begin{tabular}{c|cccc}
\toprule[1.2pt]
Metrics  &Input &DMCNN/(+SAM) &MDDM/(+SAM) &MBCNN/(+SAM)    \\
\hline 
\text{PSNR}$\uparrow$ &17.117 &19.914/\textbf{20.769} &20.088/\textbf{20.883}  &21.414/\textbf{21.532} \\

\text{SSIM}$\uparrow$ &0.5089 &0.7575/\textbf{0.7699} &0.7441/\textbf{0.7640}  &0.7932/\textbf{0.7940} \\

\text{LPIPS}$\downarrow$ &0.5314 &0.3764/\textbf{0.3630} &0.3409/\textbf{0.3299}  &0.3318/\textbf{0.3302} \\

\bottomrule[1.2pt]
\end{tabular}
}

\label{tab:SAM}
\end{table}

\subsection{More Qualitative Comparisons}
As seen in Fig. \ref{fig:uhdm_1}-\ref{fig:tip-1}, we provide more visual results and comparisons with current state-of-the-art methods on three real-world demoiréing datasets: UHDM (resolution: $3840\times 2160$), FHDMi~\cite{he2020fhde} (resolution: $1920\times 1080$) and TIP2018~\cite{sun2018moire} (resolution: $256\times 256$). Apparently, our model can remove moiré patterns more cleanly and preserve high-frequency details better.

\section{Revisit Current Multi-Scale Schemes in Image Demoiréing}
\label{sec:revisit}
We have discussed in our main paper that a key challenge in image demoiréing is the scale variation of the moiré pattern. In this section, we conduct a more detailed analysis of multi-scale schemes in current demoiréing works.
As shown in \cref{fig:multi-scale}, we summarize these schemes into two parts: single-stage training and multi-stage training. We figure out their inefficiency and insufficiency, which limit their performance when processing ultra-high-definition images. 

\subsection{Single-Stage Training}
Most of the demoiréing works adopt a single-stage framework, \ie, given a moiré image $I_{\text{moiré}}\in\mathbb{R}^{h\times w\times 3}$, an end-to-end network $\mathbf{F}$ is trained to produce the final demoiréd image $I_{\text{demoiré}}$:
\begin{equation}
I_{\text{demoiré}} = \mathbf{F}(I_{\text{moiré}})
\end{equation}
Specifically, they embed different multi-scale schemes into their networks, which can be simplified and summarized into two topological architectures: parallel multi-scale and cascaded multi-scale.

\begin{figure}[H]
   \begin{center}
      \includegraphics[width=1\textwidth]{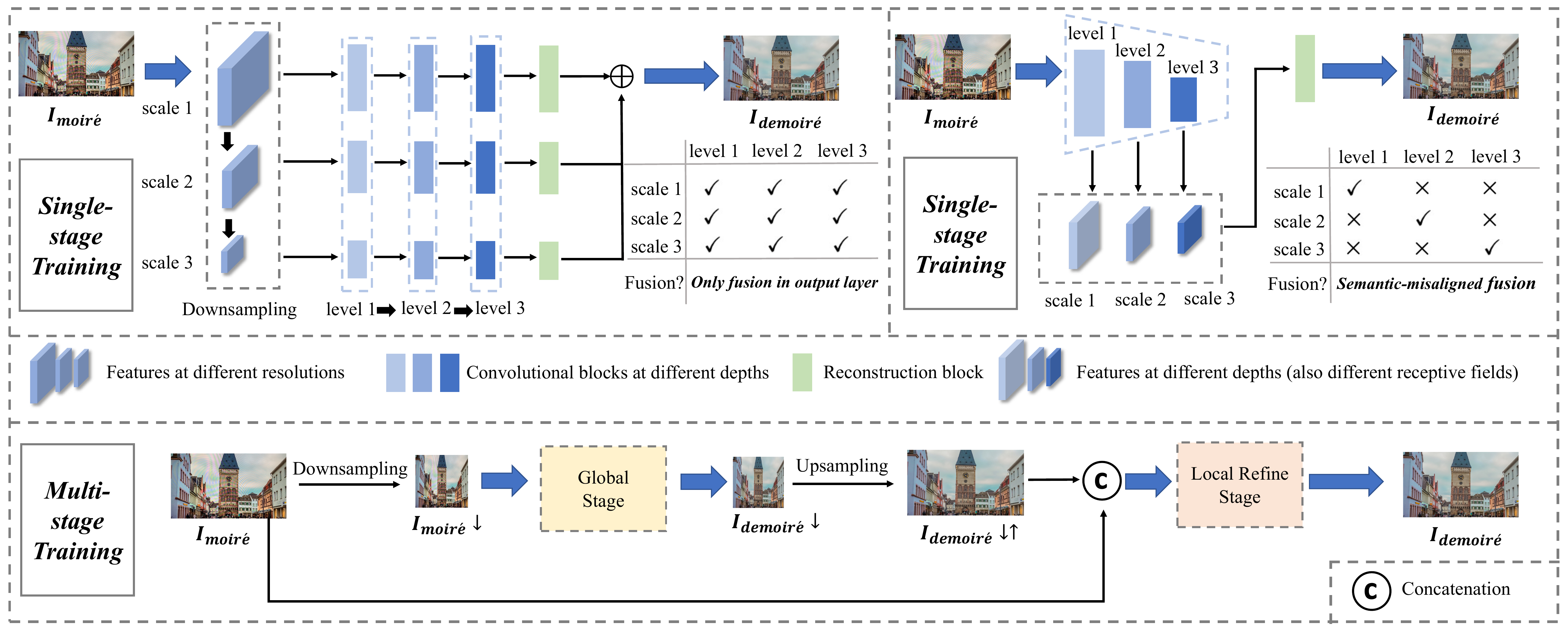}
   \end{center}
      
      \caption{A summary of current works for solving the multi-scale challenge in image demoiréing}
      
   \label{fig:multi-scale}
\end{figure}

\noindent\textbf{Cascaded multi-scale:}
Adopted by MopNet\cite{he2019mop}, MBCNN\cite{zheng2020image} and WDNet\cite{liu2020wavelet} (Note that although MopNet is a multi-stage framework, it harvests multi-scale information in one sub-network), the insight in cascaded multi-scale strategy is utilizing features from different-depth layers to get multi-scale representations. As shown in the right upper part of \cref{fig:multi-scale}, the moiré image first goes through an encoder that contains three levels to extract features. Then the intermediate results in each level are fused together and fed to the decoder for reconstruction. Since features are produced in different-depth layers, their receptive fields are different (the receptive field is larger for a deeper feature). However, another fact is ignored: features at different depths have different semantic meanings. For example, features extracted in the early layer usually contain low-level information such as edge, while features in the deeper layers contain more abstract attributes learned by the network. Recall that the scale-variation challenge means that the observed object remains the same for all attributes (\eg,  color, shape) except for the scale that appeared in an image (\ie,  pixels it counts). Thus, a more reasonable design is the network can extract multi-scale information at the same semantic level (\ie, depth level). Further, a robust network should harvest multi-scale information at each semantic level to handle different attributes. Based on this analysis, we find that this cascaded strategy lacks multi-scale ability at a specific semantic level, limiting its scale-robust ability.

\noindent\textbf{Parallel multi-scale:}
The parallel multi-scale indicates construction of parallel high-resolution to low-resolution branches to process different-scale features, as adopted in DMCNN\cite{sun2018moire} and MDDM\cite{cheng2019multi}. At each scale, several convolutional blocks are stacked to extract features and finally produce a three-channel output. Without loss of generality, we suppose there are three scales and three convolutional blocks in each scale to illustrate and analyze this strategy. 

As shown in the left upper part of \cref{fig:multi-scale}, the moiré image first goes through several downsampling convolutional heads with different strides to obtain shallow representations with different resolutions:
\begin{equation}
J_i = \text{Conv$_i$}(I_{\text{moiré}}), i=1, 2, 3
\end{equation}
where \text{Conv$_i$} denotes convolutional block with stride $s=2^{i-1}$, $J_i\in\mathbb{R}^{\frac{h}{2^{i-1}} \times \frac{w}{2^{i-1}}\times c}$. After that, each $J_i$ is fed to several convolutional blocks in parallel:
\begin{equation}
X_i = F_i^3(F_i^2(F_i^1(J_i))), i=1, 2, 3
\end{equation}
where $F_i^j$ denotes the $j$-th blocks in $i$-th scale (branch), $X_i\in\mathbb{R}^{\frac{h}{2^{i-1}} \times \frac{w}{2^{i-1}}\times 3}$. Then an upsampling layer would be utilized to align the spatial size of each-scale outputs, followed by a summation operation to get the final prediction $I_{\text{demoiré}}$:
\begin{equation}
I_{\text{demoiré}} = X_1 + X_{2\uparrow} + X_{3\uparrow\uparrow}
\end{equation}

\begin{figure*}
  \centering
    \includegraphics[width=1\linewidth]{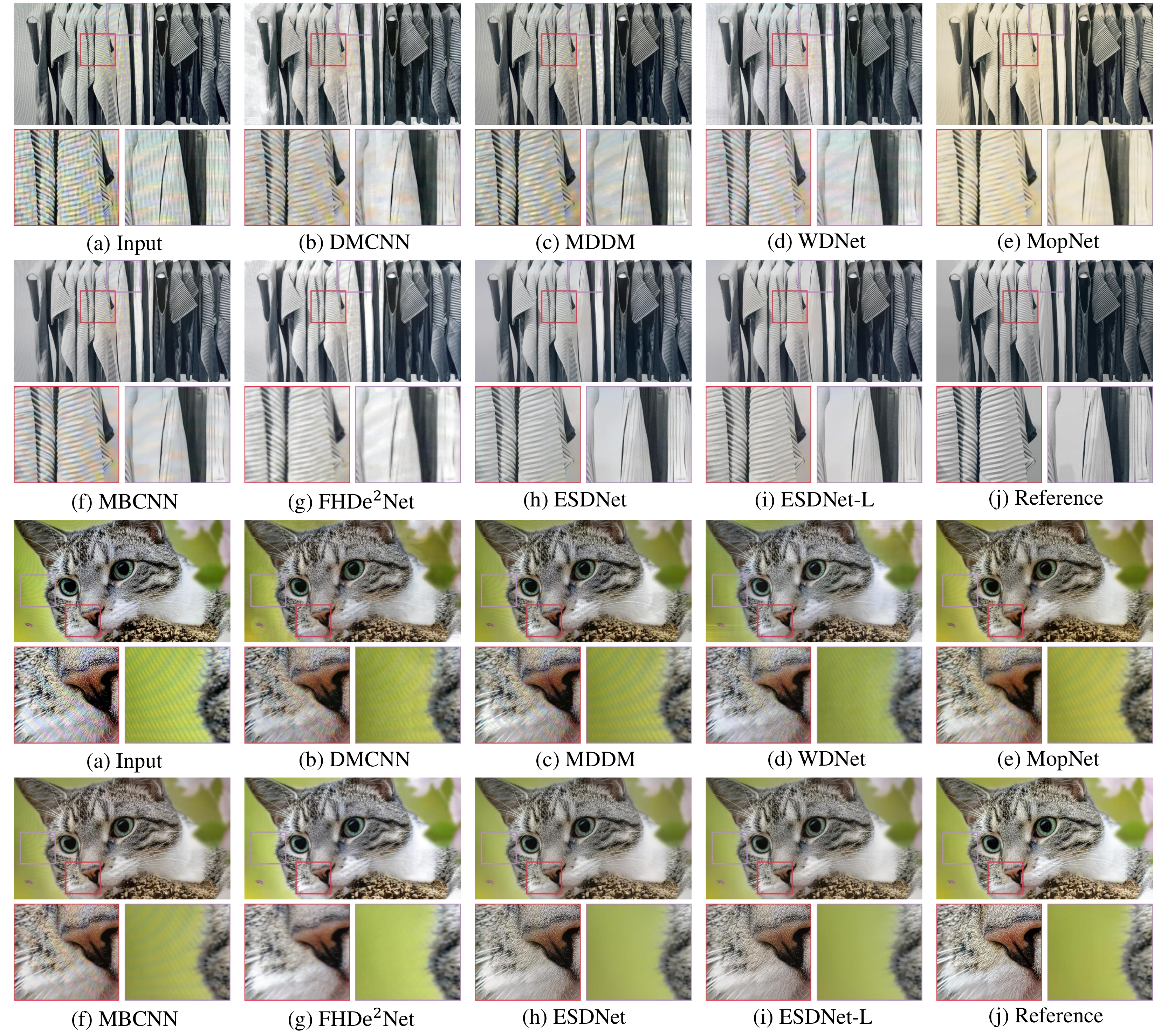}
    \caption{Qualitative comparisons of our models with other state-of-the-art methods on the UHDM dataset, ESDNet is our standard model and ESDNet-L is our larger model}
    \label{fig:uhdm_1}
\end{figure*}

\begin{figure*}
  \centering
    \includegraphics[width=1\linewidth]{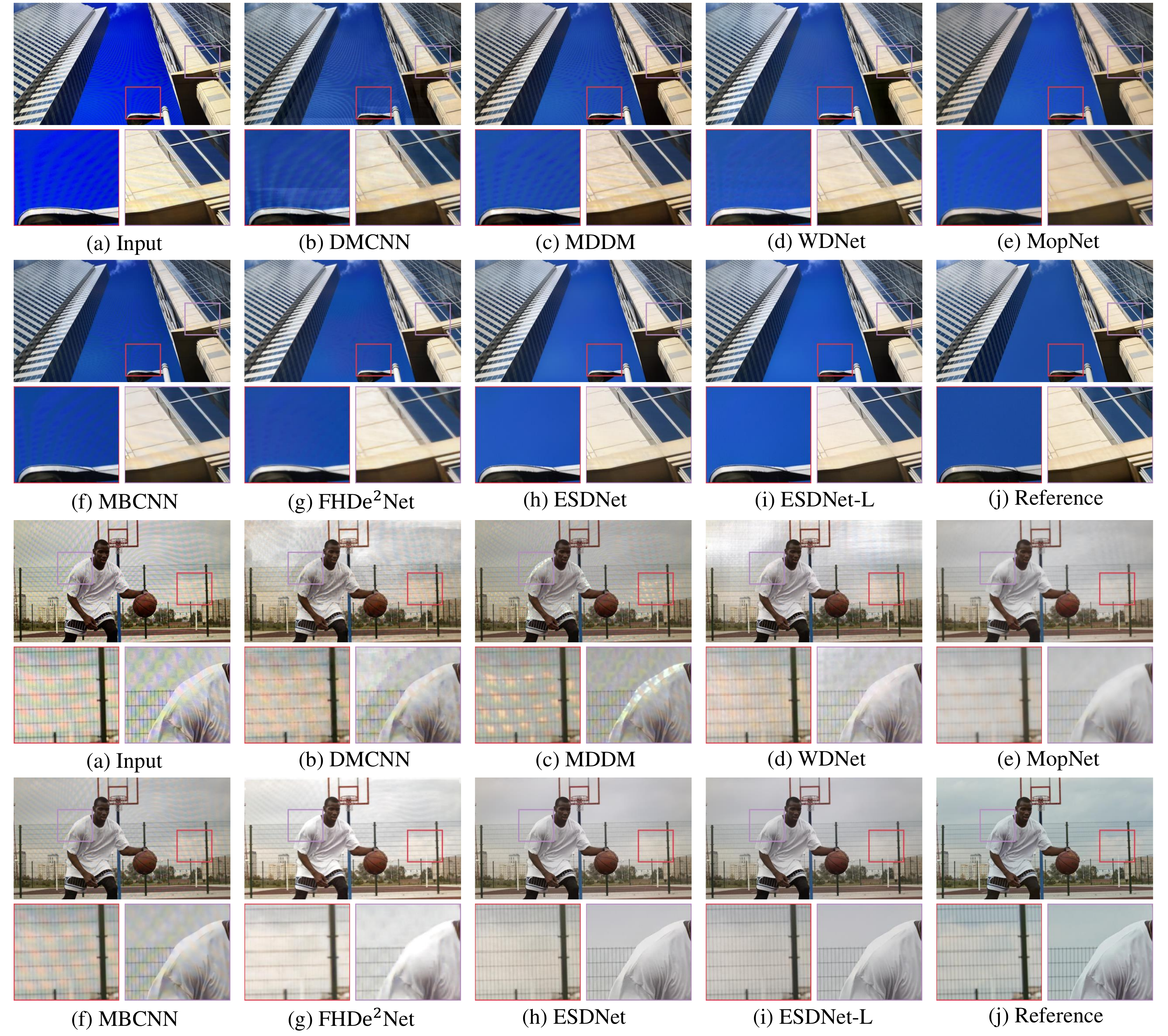}
    \caption{Qualitative comparisons of our models with other state-of-the-art methods on the UHDM dataset, ESDNet is our standard model and ESDNet-L is our larger model}
    \label{fig:uhdm_2}
\end{figure*}

\begin{figure*}
  \centering
    \includegraphics[width=1\linewidth]{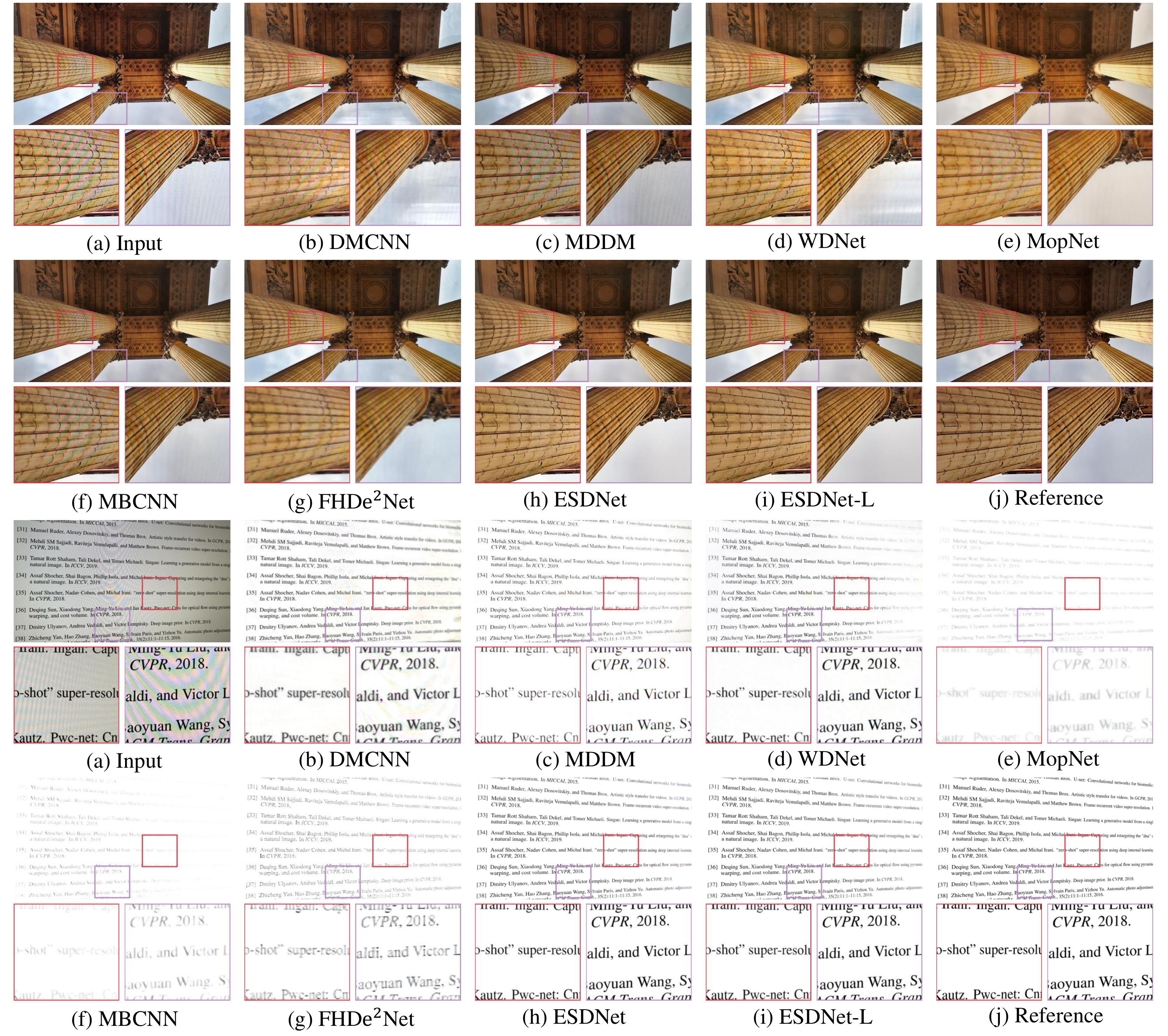}
    \caption{Qualitative comparisons of our models with other state-of-the-art methods on the UHDM dataset, ESDNet is our standard model and ESDNet-L is our larger model}
    \label{fig:uhdm_3}
\end{figure*}

\begin{figure*}
  \centering
    \includegraphics[width=1\linewidth]{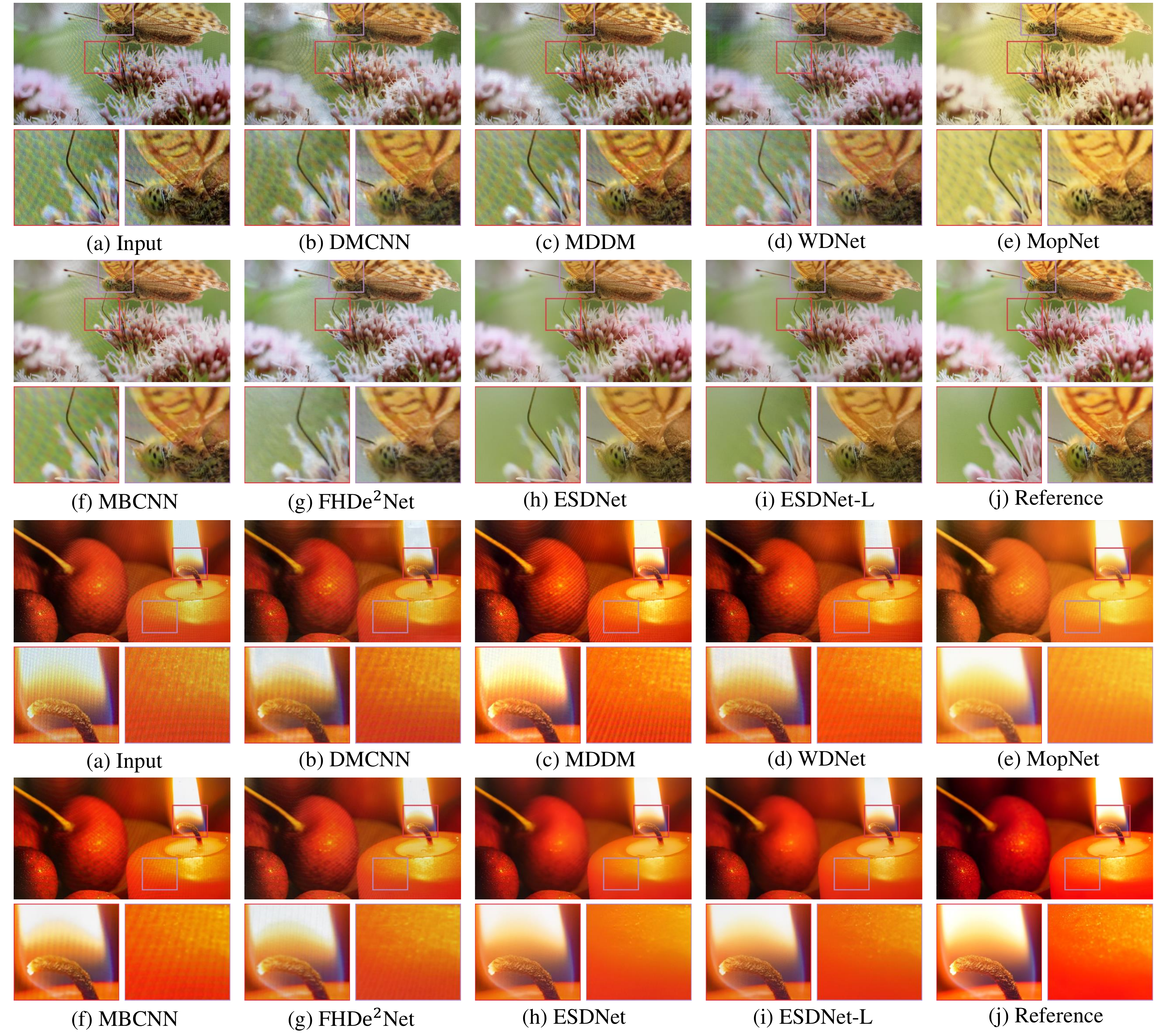}
    \caption{Qualitative comparisons of our models with other state-of-the-art methods on the UHDM dataset, ESDNet is our standard model and ESDNet-L is our larger model}
    \label{fig:uhdm_4}
\end{figure*}

\begin{figure*}
  \centering
    \includegraphics[width=1\linewidth]{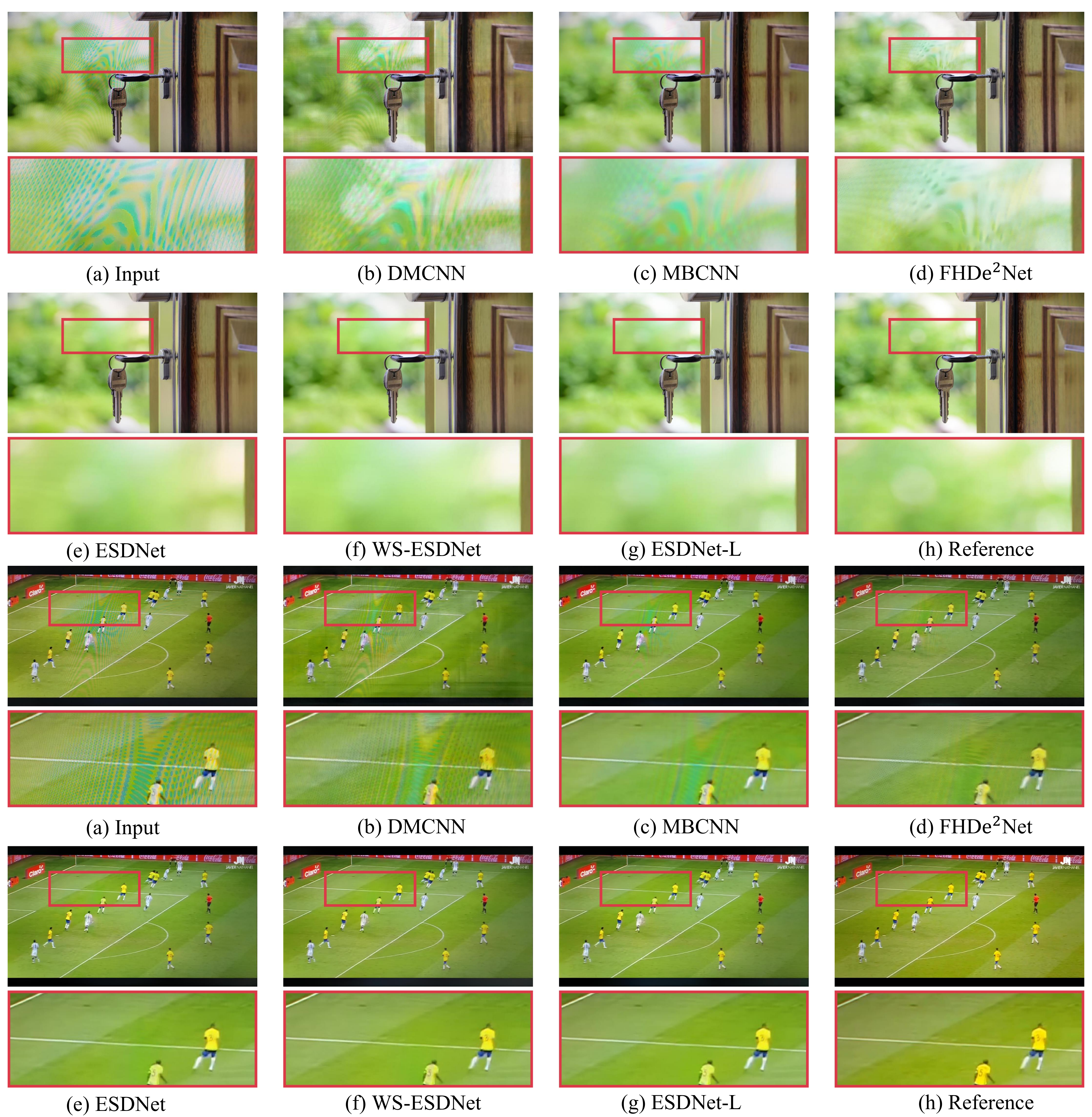}
    \caption{Qualitative comparisons of our models with three representative state-of-the-art methods on the FHDMi dataset~\cite{he2020fhde}, including DMCNN~\cite{sun2018moire}, MBCNN~\cite{zheng2020image} and FHDe$^2$Net~\cite{he2020fhde}. ESDNet is our standard model and ESDNet-L is our larger model. WS-ESDNet is our more lightweight model, the parameters of which are shared in three branches of pyramid context extraction module}
    \label{fig:short-1}
\end{figure*}

\begin{figure*}
  \centering
    \includegraphics[width=1\linewidth]{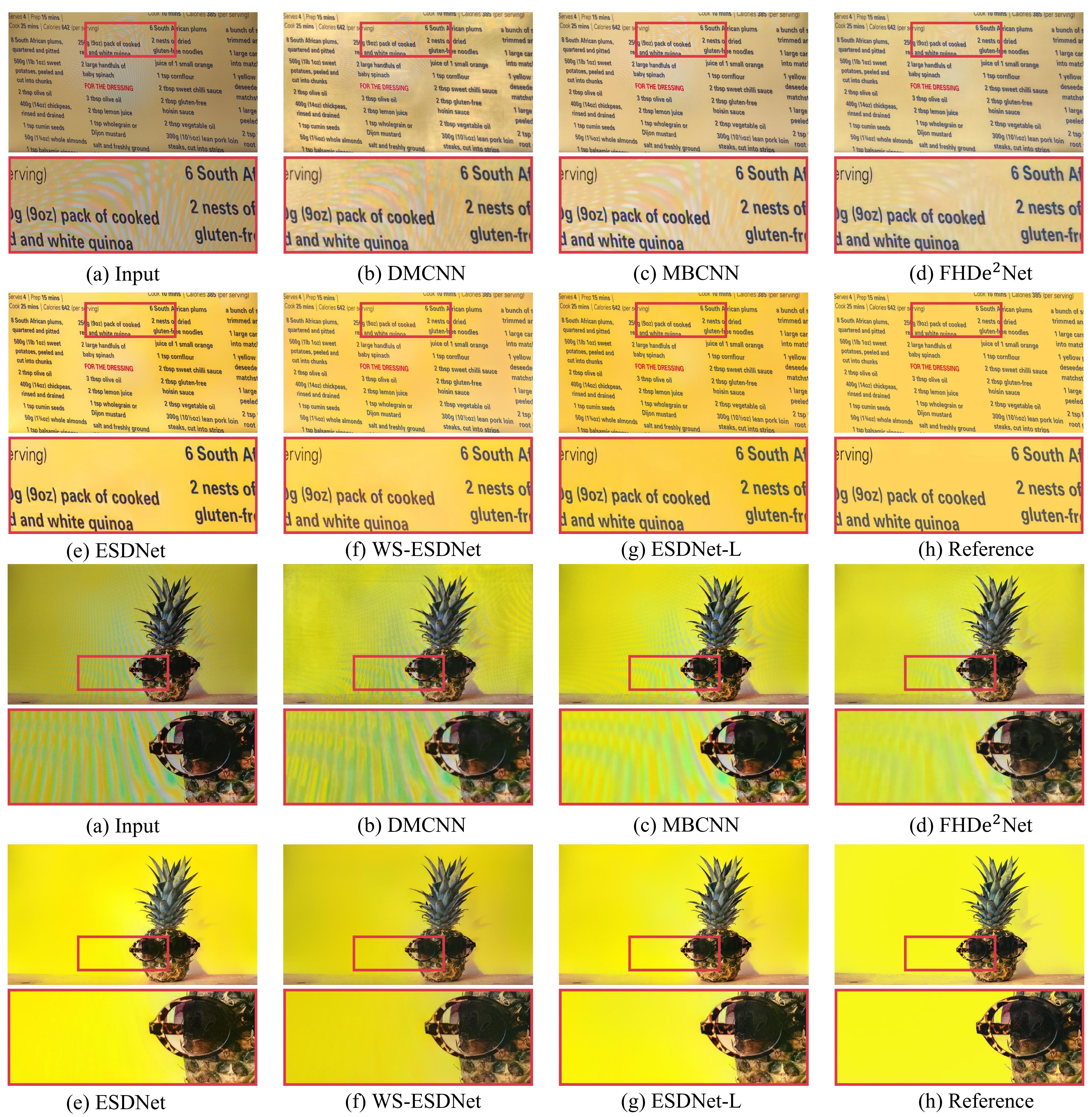}
    \caption{Qualitative comparisons of our models with three representative state-of-the-art methods on the FHDMi dataset~\cite{he2020fhde}, including DMCNN~\cite{he2020fhde}, MBCNN~\cite{zheng2020image} and FHDe$^2$Net~\cite{he2020fhde}. ESDNet is our standard model and ESDNet-L is our larger model. WS-ESDNet is our more lightweight model, the parameters of which are shared in three branches of pyramid context extraction module}
    \label{fig:short-2}
\end{figure*}

\begin{figure*}
  \centering
    \includegraphics[width=1\linewidth]{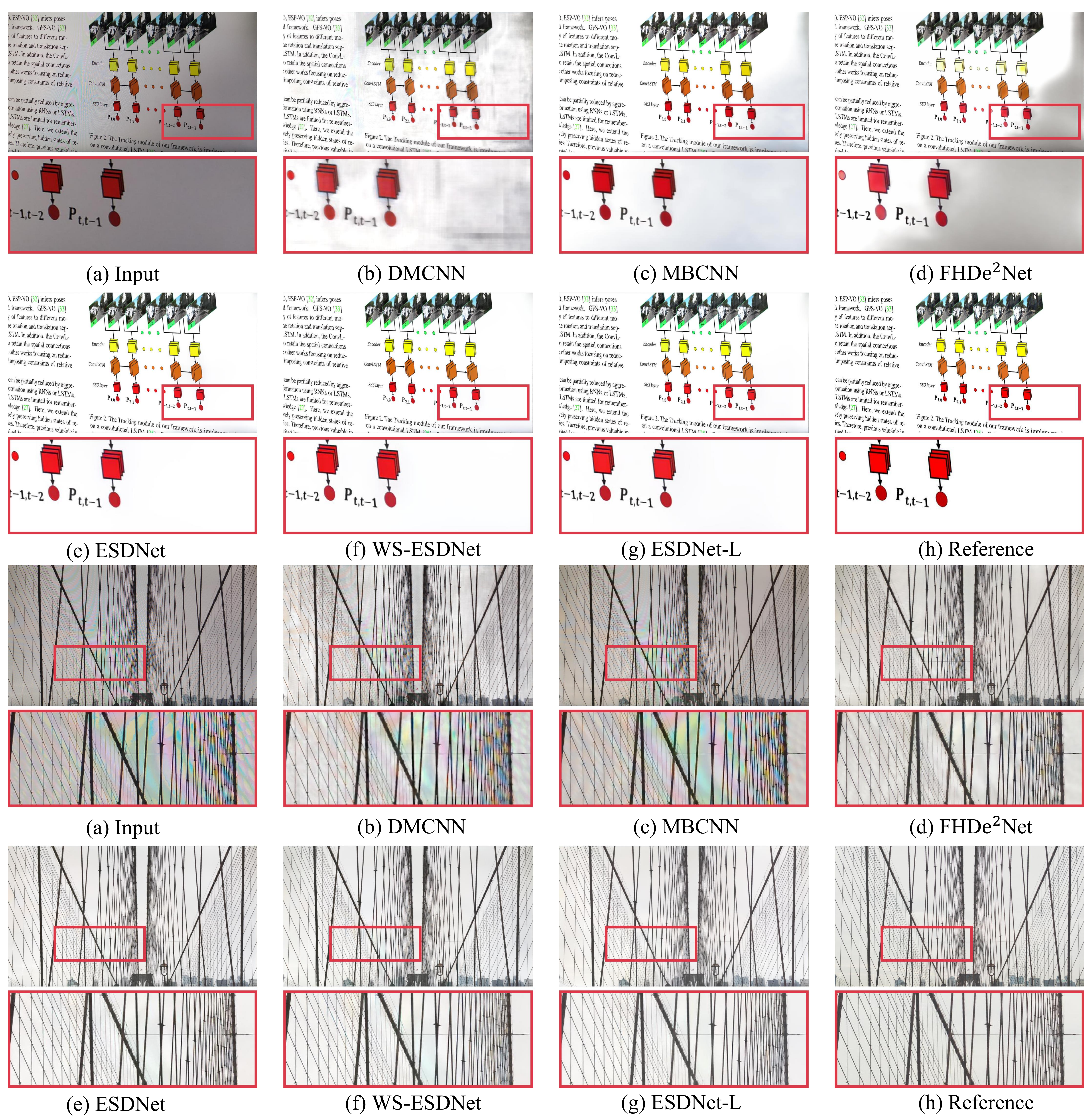}
    \caption{Qualitative comparisons of our models with three representative state-of-the-art methods on the FHDMi dataset~\cite{he2020fhde}, including DMCNN~\cite{sun2018moire}, MBCNN~\cite{zheng2020image} and FHDe$^2$Net~\cite{he2020fhde}. ESDNet is our standard model and ESDNet-L is our larger model. WS-ESDNet is our more lightweight model, the parameters of which are shared in three branches of pyramid context extraction module}
    \label{fig:short-3}
\end{figure*}

\begin{figure*}
  \centering
    \includegraphics[width=1\linewidth]{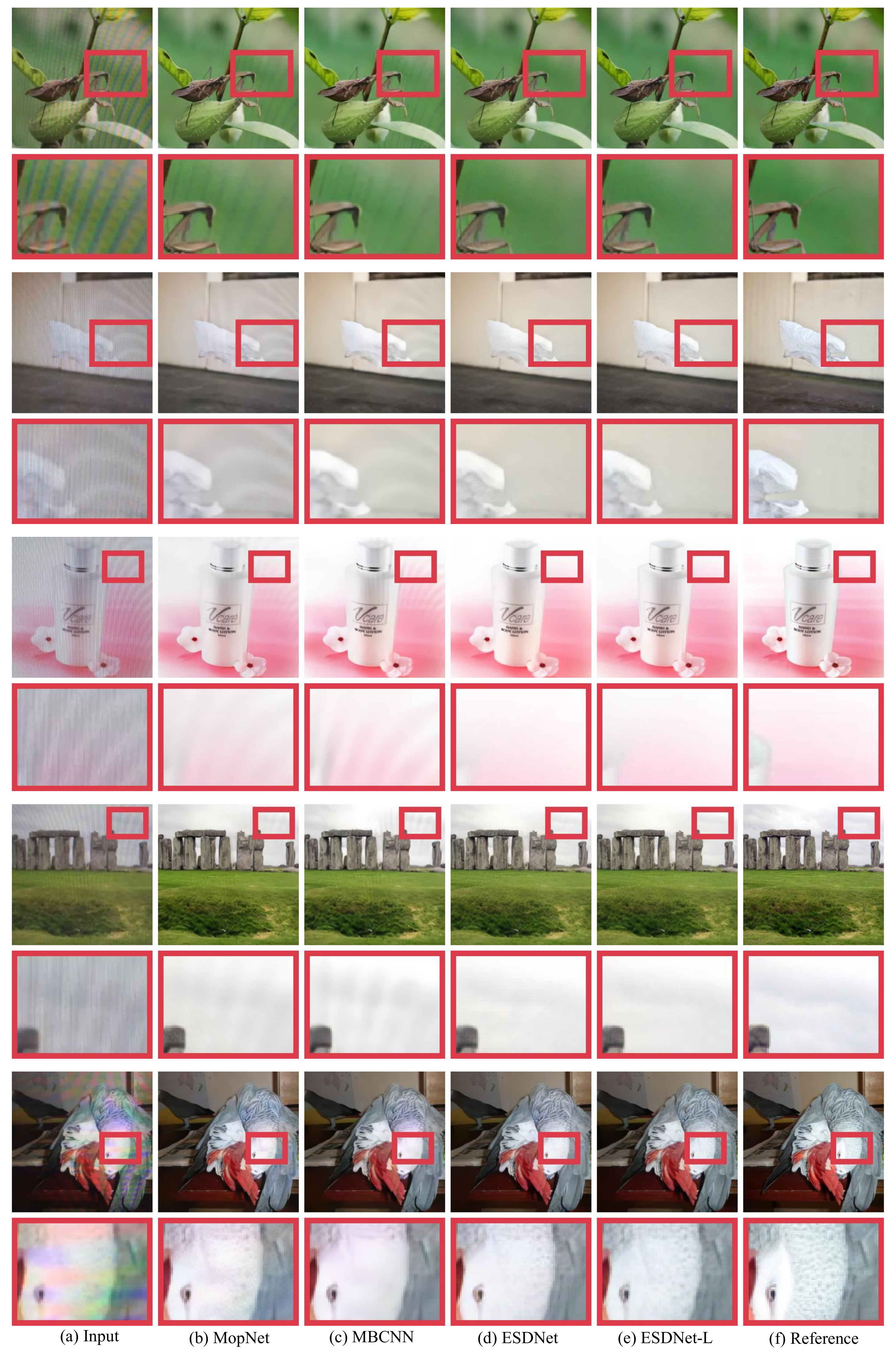}
    \caption{Qualitative comparisons of our models with two representative state-of-the-art methods on the TIP2018 dataset~\cite{sun2018moire}, including MopNet~\cite{he2019mop} and MBCNN~\cite{zheng2020image}. ESDNet is our standard model and ESDNet-L is our larger model}
    \label{fig:tip-1}
\end{figure*}

Unlike the cascaded multi-scale scheme, the insight here is to reduce the resolution at the input stage, so different branches have different receptive fields. However, the problem is, this framework only fuses the results at the end of each branch, ignoring the interaction of the intermediate features. As a result, each extracted feature is only determined by its current branch (scale), dramatically limiting the network's representation ability. For example, to produce the feature $F_2^2$, the network only utilizes the information from $F_2^1$. However, a more representative feature needs to harvest multi-scale information from last semantic level. Only fusing information in the last layer results in coarse moiré pattern removal, as shown in Fig. \ref{fig:uhdm_1}-\ref{fig:tip-1}.

\subsection{Multi-Stage Training}
FHDe$^2$Net~\cite{he2020fhde} is the only current work which proposes to tackle real-world high-definition moiré images. Due to the increased resolution, the scale of the moiré pattern would expand extremely larger, which has been the main challenge in the high-definition demoiréing. The central insight in this work is adopting a multi-stage framework to handle this problem, the networks of which are trained step by step. As shown in the lower part of \cref{fig:multi-scale}, the overall framework can be divided into two stages: the global stage and the local refine stage (In fact, it consists of four sub-networks, but we summarize it into two stages here for analysis). The input of the global stage is a downsampled low-resolution ($384\times 384$) moiré image, so the network in this stage can obtain a full-image-size receptive field. Although the large-scale moiré pattern can be removed, the images' high-frequency details are severely lost due to the downsampling operation. Hence, in the local refinement stage, the original high-resolution image would be utilized to guide the low-resolution demoiréd image to recover the details. However, our experiments find it hard for the network to differentiate the moiré pattern from the image textures, leading to the reintroduction of the moiré pattern and unsatisfactory texture recovery. Furthermore, its internal complex module design shows a heavy computational burden, which is unacceptable for ultra-high-definition image demoiréing.

\end{document}